\documentclass[11pt]{article}

\usepackage[final]{acl}

\usepackage{times}
\usepackage{fancyvrb}
\usepackage{fvextra}
\usepackage{latexsym}
\usepackage{amsmath}
\usepackage[T1]{fontenc}

\usepackage[utf8]{inputenc}

\usepackage{booktabs}
\usepackage{graphicx}
\usepackage{placeins}
\usepackage{multirow}
\usepackage{array}
\usepackage[table]{xcolor}  
\usepackage{enumitem}
\usepackage{pifont}
\usepackage{algorithm}
\usepackage{algpseudocode}
\newcommand{\cmark}{\textcolor{green!45!black}{\ding{51}}}
\newcommand{\xmark}{\textcolor{red!70!black}{\ding{55}}}

\usepackage{microtype}

\usepackage{inconsolata}

\usepackage{hyperref}
\hypersetup{
    colorlinks=true,
    linkcolor=blue,
    citecolor=blue,
    urlcolor=blue
}
\title{RetailBench: Benchmarking long horizon reasoning and coherent decision making of LLM agents in realistic retail environments}

\author{
Linghua Zhang\textsuperscript{1}
\quad
Jun Wang\textsuperscript{2}
\quad
Jingtong Wu\textsuperscript{2}
\quad
Zhisong Zhang\textsuperscript{\dag} \\
\textsuperscript{1}Ant Group
\quad
\textsuperscript{2}City University of Hong Kong \\
\texttt{zlh20011228@gmail.com, nanrong.wj@ant-intl.com} \\
\texttt{jingtong.wujt@ant-intl.com, zhisong.zhang@cityu.edu.hk}
}

\begin{document}
\maketitle
\begin{abstract}
Large language model (LLM) agents have made rapid progress on short-horizon, well-scoped tasks, yet their ability to sustain coherent decisions in dynamic long-horizon environments remains uncertain. We introduce RetailBench, a data-grounded simulation benchmark for evaluating tool-using LLM agents in single-store supermarket operation. RetailBench models retail management as a partially observable decision process and is designed to support thousand-day-scale simulations. In this environment, agents must manage pricing, replenishment, supplier selection, shelf assortment, inventory aging, customer feedback, external events, and cash-flow constraints. We evaluate seven contemporary LLMs under representative agent frameworks over a 180-day evaluation horizon and compare them with a privileged oracle policy. Results show substantial variation across models: only a small subset survives the full evaluation horizon, and even the strongest LLM runs remain substantially behind the oracle policy in final net worth and sales outcomes. Behavioral analysis attributes these gaps to incomplete evidence acquisition, surface-level decision making, and the lack of a consistent long-horizon policy. RetailBench provides a controlled testbed for studying reliable autonomy in economically grounded long-horizon decision-making.

\end{abstract}

\section{Introduction}
\label{sec:intro}
\begin{figure}[t]
\centering
\includegraphics[width=\columnwidth]{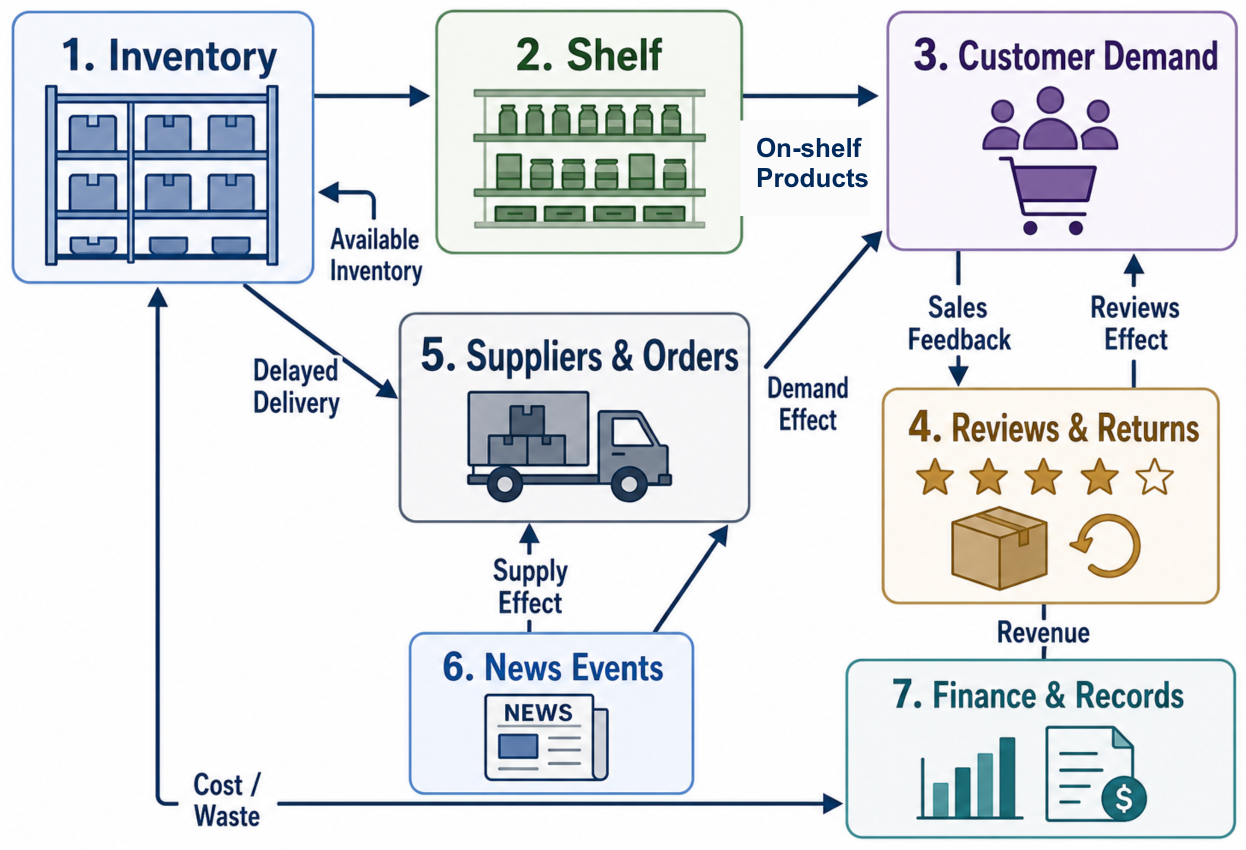}
\caption{Overview of the RetailBench simulation environment. Inventory, shelf assortment, customer demand, suppliers and orders, reviews and returns, news events, and finance and records are coupled in this environment.}
\label{fig:retailbench_environment}
\vspace{-2mm}
\end{figure}

Large language models (LLMs), particularly when equipped with reasoning and tool-use capabilities, have achieved strong performance on short-horizon, structured tasks such as code editing, mathematical problem solving, complex information retrieval, and web interaction \citep{swe-bench, hle, omni-olpc, mialon2023gaiabenchmarkgeneralai, browsecomp, webarenarea, mind2web, tbench_2025}. As these benchmarks become increasingly saturated, the central question in agent evaluation is shifting from isolated task completion to sustained autonomy: can LLM agents pursue persistent objectives, adapt to evolving feedback, and recover from imperfect prior decisions over extended horizons?

Existing evidence suggests that this remains a major challenge. Recent studies show that even state-of-the-art agents often fail to maintain coherent strategies across long interactions, especially when local errors compound through delayed consequences and reshape future state distributions \citep{amodei2024machines, kwa2025measuringaiabilitycomplete, metr2025measuring, nof1, andonlabs2025vendingbench2, backlund2025vendingbenchbenchmarklongtermcoherence}. In response, benchmarks such as UltraHorizon, OdysseyBench, and HeroBench have begun to evaluate agents in temporally extended and more realistic environments \citep{utltabench,odysseybenchevaluatingllmagents,herobenchbenchmarklonghorizonplanning}. These efforts point to a broader methodological need: agent benchmarks should move beyond isolated task completion and evaluate whether agents can maintain persistent objective alignment and stable behavioral consistency in realistic long-horizon settings.

Retail business operation provides a natural testbed for this form of long-horizon agency. A store operator must repeatedly monitor inventory and cash flow, interpret historical sales and customer feedback, forecast future demand, and decide pricing, replenishment, and assortment actions whose effects may emerge only after several days. Prior work has studied related components, including retail and promotion simulation \citep{xia2023retailsynth,xia2024simulationretailpromotions}, transaction and demand modeling \citep{wang2025freshretailnet,tkachuk2024consumertransactions}, online shopping \citep{wang2025shoppingbench,3webshopscalablerealworldweb}, and inventory or order-fulfillment control \citep{OFCOURSE,inventory,leluc2023marlimmultiagentreinforcementlearning,invagentlargelanguagemodel,aimbenchevaluatingdecisionmakingbiases}. Yet these benchmarks typically isolate individual subproblems and do not capture the coupled, closed-loop dynamics of real business operation. As a result, the community lacks an open, real-data-grounded benchmark for evaluating tool-using LLM agents under diverse operational pressures over extended horizons.

We introduce \textit{RetailBench}, a controlled simulator for single-store supermarket operation grounded in real retail product and sales signals. In \textit{RetailBench}, agents repeatedly make pricing, replenishment, supplier-selection, information-acquisition, memory-management, and day-termination decisions while facing delayed deliveries, inventory aging, stockouts, customer reviews, returns, and financial constraints.

Our contributions are threefold:
\begin{itemize}[leftmargin=*]
    \item \textbf{A long-horizon retail environment.} We introduce \textit{RetailBench}, a data-grounded benchmark that evaluates LLM agents under coupled pricing, inventory, supplier, feedback, and financial dynamics.
    \item \textbf{An evaluation of current agent systems.} We benchmark representative agent frameworks and state-of-the-art LLMs, showing that current agents remain far from stable, high-performing long-horizon operational policies.
    \item \textbf{A systematic failure analysis.} We identify three recurring failure modes: \textbf{incomplete evidence acquisition}, \textbf{surface-level decision making}, and \textbf{lack of a consistent long-horizon policy}.
\end{itemize}

\section{Environment Construction}
\label{sec:env}

\begin{figure*}[t]
\centering
\includegraphics[width=0.92\textwidth]{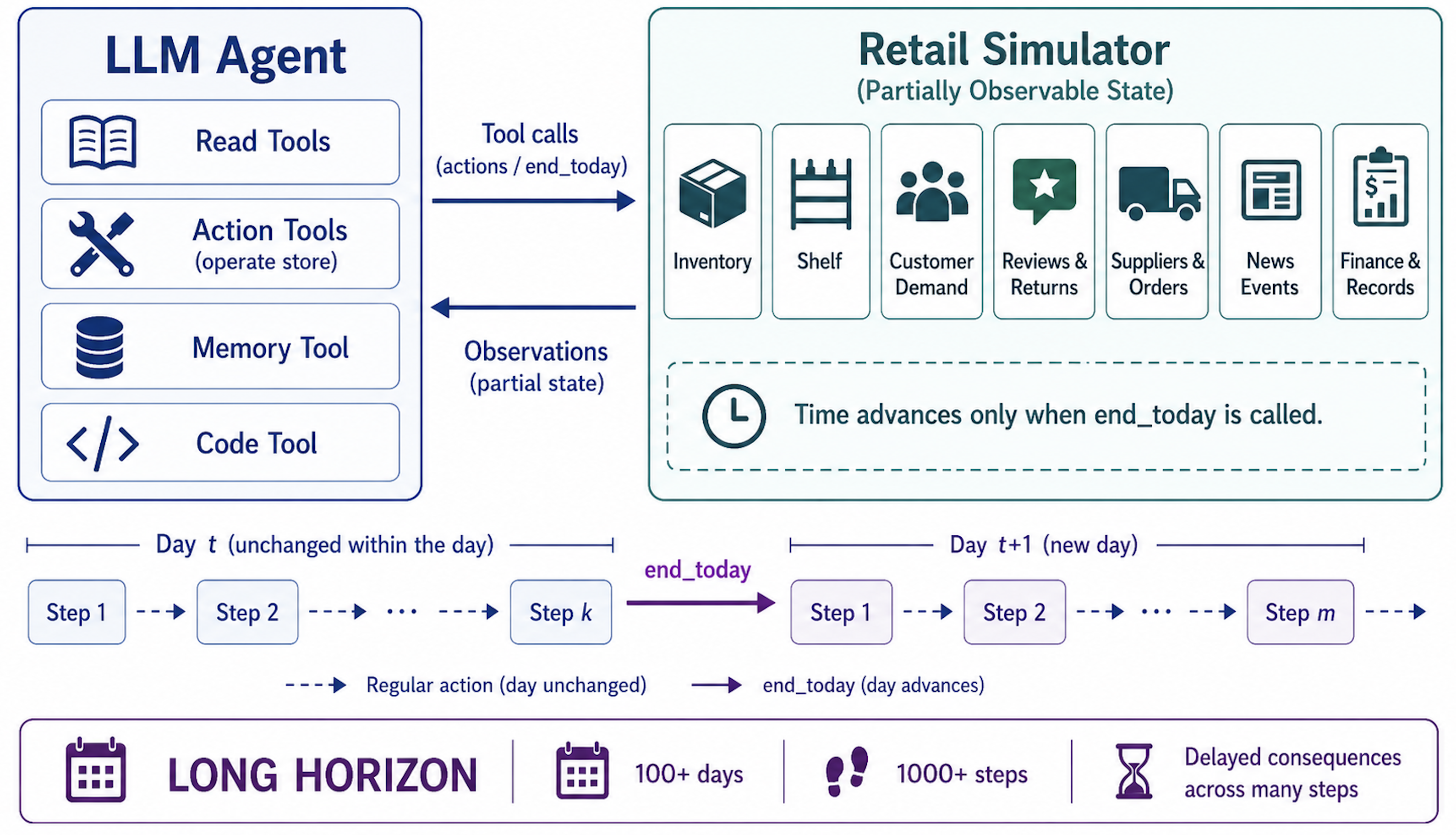}
\caption{RetailBench interaction framework. A tool-using LLM agent operates a simulated retail store through read, action, memory, and code tools. The simulator state is partially observable, and time advances from day $t$ to day $t+1$ only when the agent invokes the end-today action.}
\label{fig:retailbench_framework}
\vspace{-3mm}
\end{figure*}

\subsection{Interface}

Figure~\ref{fig:retailbench_framework} summarizes RetailBench as an interaction between a tool-using agent and a partially observable retail simulator. Formally, each step is modeled as a POMDP:
\begin{equation}
\label{eq:pomdp}
\mathcal{M}=(\mathcal{S},\mathcal{A},\mathcal{O},T,\Omega,R)
\end{equation}
Here, $\mathcal{S}$ is the simulator state, $\mathcal{A}$ is the tool-action space, $\mathcal{O}$ is the exposed observation space, $T$ is the transition model, $\Omega$ is the observation function, and $R$ records business outcomes. Appendix~\ref{app:state_action_spaces} provides the detailed state space $\mathcal{S}$ and action space $\mathcal{A}$ definitions. This abstraction matches Figure~\ref{fig:retailbench_framework}: the agent interacts only with tool-facing observations and actions, while the simulator privately maintains delayed inventory, shelf visibility, demand, supplier, feedback, and cash-flow consequences.

The latent state on day $t$ and intra-day step $k$ is factorized as:
\begin{equation}
\label{eq:latent_state}
S_{t,k} = (P_{t,k}, I_{t,k}, V_{t,k}, C_{t,k}, D_{t,k}, E_{t,k}, F_{t,k})
\end{equation}
Here, $P_{t,k}$ stores product attributes and prices, $I_{t,k}$ stores inventory and pending orders, $V_{t,k}$ stores the active shelf assortment, $C_{t,k}$ stores supplier candidates and conditions, $D_{t,k}$ stores demand factors, $E_{t,k}$ stores external news, and $F_{t,k}$ stores financial accounting. These factors instantiate a single-store supermarket where pricing, replenishment, shelf placement, supplier choice, customer feedback, and cash-flow pressure interact over multiple days.

At each step, the agent selects a tool action from three tool groups:
\begin{equation}
\label{eq:agent_interaction}
\begin{aligned}
a_{t,k} &\in \mathcal{A},\\
\mathcal{A} &= \mathcal{A}^{\mathrm{read}}
\cup \mathcal{A}^{\mathrm{other}}
\cup \mathcal{A}^{\mathrm{act}},\\
o_{t,k} &= \Omega(S_{t,k},a_{t,k}), \quad a_{t,k}\in\mathcal{A}^{\mathrm{read}}\\
z_{t,k} &= G(o_{\le t,k},a_{t,k}), \quad a_{t,k}\in\mathcal{A}^{\mathrm{other}}\\
S_{t,k+1} &= T(S_{t,k},a_{t,k}), \quad a_{t,k}\in\mathcal{A}^{\mathrm{act}}
\end{aligned}
\end{equation}
The read-only subset $\mathcal{A}^{\mathrm{read}}$ provides access to tools for querying funds, inventory levels, shelf status, product prices, inventory costs, sales and profit histories, supplier quotes and price histories, notes, and, when enabled, review ratings, supplier return rates, and news summaries. These tools expose observations $o_{t,k}$ only; they do not disclose latent utilities, hidden news-impact parameters, raw supplier-quality scores, or any other privileged simulator state.

The auxiliary subset $\mathcal{A}^{\mathrm{other}}$ contains constrained analysis tools, the \texttt{memory} tool, and \texttt{execute\_code}. The \texttt{memory} tool allows the agent to write long-form memory entries, whereas \texttt{execute\_code} can aggregate observations and call read-only tools inside a sandboxed environment. Neither tool can modify prices, orders, shelf placement, or inventory.

The state-changing subset $\mathcal{A}^{\mathrm{act}}$ comprises product price updates, replenishment orders, shelf-assortment updates, cross-day note writing, and day termination. Once the agent invokes \texttt{end\_today}, the simulator applies the following day-level transition:
\begin{equation}
\label{eq:end_today_transition}
S_{t+1,0}=T(S_{t,k},\mathtt{end\_today})\equiv T_{\mathrm{day}}(S_{t,k}).
\end{equation}
Here, $T_{\mathrm{day}}$ denotes the day-level transition induced by \texttt{end\_today}. It processes deliveries, shelf-visible demand, sales, reviews, returns, expiration, rent, cash-flow updates, and supplier/news changes. The agent's goal is to sustain store operation while improving sales, gross profit, net worth, and operating duration.

\subsection{Data-Grounded Simulator}

RetailBench is data-grounded in three complementary ways. First, real grocery retail records define the product universe, category structure, historical prices, sales trajectories, traffic patterns, and procurement-cost priors, providing empirical priors for demand, pricing, and sourcing behavior. Second, dynamic feedback channels are grounded in external real-world corpora: review texts are sampled from category-specific review pools, and news events are adapted from financial news articles. Third, the simulator separates observable business-facing records from hidden state, using latent metadata to drive demand, returns, supplier costs, and cash-flow consequences.

The primary retail source is the Dominick's dataset~\citep{DominicksDataset}, which provides store-level grocery records on product categories, product-level sales, prices, and cost-related signals. We use these records to define the product universe, initialize prices, estimate demand and traffic patterns, and construct procurement-cost priors. From the 20 available categories, we select 96 products with sufficiently complete records for consistent estimation of price, sales, and category-level substitution patterns. Thus, product availability, baseline demand, price sensitivity, traffic scale, and cost assumptions are anchored in observed retail behavior rather than manually specified product profiles.

Reviews and news introduce controlled but corpus-grounded dynamics. Amazon Reviews 2023~\citep{hou2024bridging} provides category-specific review pools annotated with ratings, sentiment, and product-aspect dimensions. During simulation, realized sales trigger product-level reviews whose ratings depend on supplier quality, with comments sampled from the corresponding category--rating--aspect pool. The financial-news-articles dataset~\citep{ashraq2025financialNewsArticles} is used to construct neutral, category-level, and product-level news events. Agents observe only event titles, content, and historical records, while hidden metadata determines each event's demand- or supply-side effects. RetailBench should therefore be interpreted as a reproducible, data-grounded stress test for long-horizon agent behavior, rather than as a full-fidelity simulator for forecasting real-store financial outcomes.

\subsection{Modules}
\label{sec:module_logic}
The simulator consists of the seven modules shown in Figure~\ref{fig:retailbench_environment}: Inventory, Suppliers \& Orders, Shelf, Customer Demand, Reviews \& Returns, News Events, and Finance \& Records. These modules share static product attributes and are coupled through daily transitions that jointly update delayed deliveries, shelf visibility, sales, feedback, news effects, inventory aging, and cash-flow accounting. Appendix~\ref{app:simulator_modules} and Appendix~\ref{app:simulator_update_equations} provide the detailed module definitions, state variables, and day-level update equations.

\subsection{Scope}
RetailBench evaluates long-horizon, tool-using LLM agents in single-store supermarket operation. Prior work has studied retail simulation and promotion optimization~\citep{xia2023retailsynth,xia2024simulationretailpromotions}, transaction and demand modeling~\citep{wang2025freshretailnet,tkachuk2024consumertransactions}, web-based shopping agents~\citep{wang2025shoppingbench,3webshopscalablerealworldweb}, supply-chain and inventory control~\citep{OFCOURSE,inventory,leluc2023marlimmultiagentreinforcementlearning,invagentlargelanguagemodel,aimbenchevaluatingdecisionmakingbiases}, and long-running business-agent evaluation~\citep{backlund2025vendingbenchbenchmarklongtermcoherence,andonlabs2025vendingbench2}. These benchmarks are valuable, but they typically expose narrower decision surfaces or isolate individual operational subproblems. RetailBench instead places agents in a closed-loop, data-grounded environment that requires coordinated decisions across pricing, inventory, suppliers, shelf placement, customer feedback, external events, and cash-flow survival. Appendix Table~\ref{tab:benchmark_axes} provides a compact scope comparison with representative benchmarks, but our main design motivation comes from the closed-loop operational requirements above.

\section{Evaluation Protocol}
\label{sec:evaluation_protocol}
\subsection{Environment Configuration}

All evaluations are conducted under a fixed RetailBench environment configuration. Specifically, the environment includes 96 products spanning 20 product categories, with five candidate suppliers for each product whose price--quality profiles evolve over time. It also incorporates customer review feedback and daily news events. The environment further specifies an initial budget of 30{,}000, a daily rent of 600, an inventory capacity of 15{,}000 units, and a shelf capacity of 40 products. The complete parameter settings are provided in Appendix~\ref{appendix:env_setting}.

\subsection{Models and Agent Frameworks}

We instantiate LLM agents with three prompting-based frameworks when available: ReAct~\citep{yao2023react}, Reflection~\citep{reflection}, and Plan-and-Act~\citep{erdogan2025planandactimprovingplanningagents}. These frameworks represent common tool-use patterns for LLM agents and interact with RetailBench through the same tool interface.

All evaluations use a fixed RetailBench environment configuration. We evaluate seven contemporary LLMs and report, for each model, its strongest completed run among the available framework settings. Because long-horizon agent performance varies substantially, this selected-run protocol summarizes each model by its best observed completed run rather than average framework-agnostic behavior. Due to API cost constraints, OpenAI GPT-5.5 is evaluated only under ReAct~\citep{yao2023react}. We additionally include a privileged oracle policy as a non-agent reference; Appendix~\ref{appendix:handcrafted_policy} provides the full rule specification.
\subsection{Metrics}

We group the evaluation metrics into five categories, each capturing a distinct operational dimension of agent performance in RetailBench. We use $\uparrow$ and $\downarrow$ to indicate metrics for which higher and lower values are preferred, respectively; tool-use metrics are reported as diagnostic measures rather than direct optimization objectives.

\begin{itemize}[leftmargin=*]
    \item \textbf{Operational viability}: \textbf{Final Net Worth} ($\uparrow$), the terminal asset value including cash, on-hand inventory value, and pending-order value; and \textbf{Survival Days} ($\uparrow$), the number of days the agent keeps the store operational.

    \item \textbf{Sales effectiveness}: \textbf{Final Total Sales} ($\uparrow$), the cumulative number of units sold over the evaluation horizon; \textbf{Daily Sold Products} ($\uparrow$), the average number of distinct products sold per day; and \textbf{Return Ratio} ($\downarrow$), the fraction of sold units that are returned.

    \item \textbf{Inventory control}: \textbf{Stockout Days} ($\downarrow$), the number of days on which at least one product has insufficient inventory; and \textbf{Expired Ratio} ($\downarrow$), the fraction of expired units among all units that are either sold or expired.

    \item \textbf{Tool-use behavior}: \textbf{Direct Tool Calls per Day} ($\uparrow$), the number of top-level tool calls made outside code execution, and \textbf{Total Tool Calls per Day} ($\uparrow$), which additionally includes tool calls issued through code execution. These metrics characterize the agent's reliance on direct tool interaction and code-mediated tool use.

    \item \textbf{Token use}: \textbf{Tokens per Day} ($\downarrow$), the average daily token consumption during evaluation.
\end{itemize}

\begin{table*}[t]
\centering
\footnotesize
\setlength{\tabcolsep}{2.2pt}
\renewcommand{\arraystretch}{1.08}
\resizebox{\textwidth}{!}{%
\begin{tabular}{llrrrrrrrrrr}
\toprule
Entry & Framework & Days $\uparrow$ & Networth $\uparrow$ & Sales $\uparrow$ & Sold products/d $\uparrow$ & Return $\downarrow$ & Expired $\downarrow$ & Stockout $\downarrow$ & Direct tools/d $\uparrow$ & Total tools/d $\uparrow$ & Tokens/d $\downarrow$ \\
\midrule
DeepSeek-V4-Pro & Plan-and-Act & \textbf{180} & \textbf{10,120.90} & \textbf{164,417} & \textbf{32.57} & \textbf{0.0858} & 0.0144 & 116 & 23.23 & \textbf{62.35} & 508,516 \\
GLM-5.1 & ReAct & 60 & -2,134.36 & 7,016 & 3.97 & 0.1347 & 0.0297 & 14 & \textbf{37.58} & 39.98 & 1,054,511 \\
GPT-5.5 & ReAct & \textbf{180} & \textbf{24,350.98} & \textbf{136,405} & \textbf{32.27} & \textbf{0.0646} & \textbf{0.0113} & 137 & 17.09 & \textbf{106.86} & \textbf{176,260} \\
Grok-4.3 & ReAct & 58 & 828.40 & 11,305 & 6.74 & 0.1028 & 0.0577 & \textbf{24} & 9.50 & 10.36 & \textbf{144,582} \\
Kimi-K2.6 & ReAct & 130 & 792.19 & 86,214 & 25.73 & 0.1226 & 0.0411 & 62 & \textbf{26.39} & 45.76 & 523,108 \\
MiniMax-M2.5 & Plan-and-Act & 73 & -397.95 & 23,521 & 12.08 & 0.1077 & \textbf{0.0049} & 52 & 18.67 & 40.19 & 484,698 \\
Qwen3.5-397B-A17B & Reflection & 71 & 1,183.17 & 35,622 & 26.17 & 0.1047 & 0.0655 & \textbf{48} & 19.87 & 49.76 & 686,103 \\
\midrule
Oracle Policy & Oracle & 180 & 131,510.42 & 267,998 & 34.48 & 0.0201 & 0.0059 & 122 & 141.44 & 141.44 & 0 \\
\bottomrule
\end{tabular}
}
\caption{Main selected-run results under a single RetailBench environment configuration. Arrows indicate the preferred direction, and bold values mark the top two LLM-agent results for each metric. The Framework column reports the framework selected by the survival-first rule: longest survival horizon, with final net worth and total sales as tie-breakers. The oracle-style policy is a privileged non-LLM reference rather than a directly comparable language-agent baseline.}
\label{tab:selected_main_results}
\vspace{-3mm}
\end{table*}

\section{Experiments}
\label{sec:experiment}

\subsection{Main Benchmark Results}

Table~\ref{tab:selected_main_results} reports the main results under our selected-run evaluation protocol. For each LLM, we select the best valid rollout according to a survival-first rule: we first choose the rollout with the longest survival horizon across the available agent frameworks, and use final net worth and total sales as tie-breakers when necessary. This protocol emphasizes operational viability, which is necessary for all downstream store-management objectives in RetailBench, while still reporting business outcomes for the selected run. The oracle policy is included as a privileged reference rather than a directly comparable language-agent baseline. It follows a manually specified quality-aware operating policy with access to structured simulator fields; Appendix~\ref{appendix:handcrafted_policy} provides the full rule specification.

\subsection{Performance}

The selected LLM runs exhibit substantial heterogeneity in long-horizon operational performance. Under the survival-first selection protocol, survival ranges from 58 to 180 days, with an average of roughly 107 days. DeepSeek-V4-Pro and GPT-5.5 complete the full 180-day horizon, Kimi-K2.6 survives for 130 days, and the remaining selected runs terminate after 58 to 73 days. These results show that long-term operational viability remains difficult for most LLM agents. Among the two full-horizon LLM runs, GPT-5.5 is more token-efficient, using roughly one-third of DeepSeek-V4-Pro's token budget; apart from DeepSeek-V4-Pro, most open-source models terminate substantially earlier.

Full-horizon survival, however, does not imply effective retail operation. The oracle-style policy reaches a final net worth of 131{,}510.42 and total sales of 267{,}998, substantially outperforming the strongest LLM runs. GPT-5.5 achieves the highest selected LLM net worth, 24{,}350.98, while DeepSeek-V4-Pro achieves the highest selected LLM sales volume, 164{,}417. Relative to the oracle-style policy, these best LLM results fall short by 107{,}159.44 in net worth and 103{,}581 in sales, respectively, and also show clear gaps in return and expiration rates. These gaps suggest that current LLM agents may sustain operation over long horizons, but still struggle to convert operational continuity into effective pricing, sourcing, inventory control, and sales outcomes.

\section{Analysis}
\label{sec:failure_pattern_analysis}
\label{sec:failure_analysis}

To understand systematic LLM failures in RetailBench, we analyze agent behavior through four operational diagnostics: \textit{product candidate selection}, \textit{evidence acquisition}, \textit{action conversion}, and \textit{temporal follow-up}. Together, these diagnostics capture the core operational loop of retail management: selecting products to manage, acquiring decision-relevant evidence, translating evidence into actions, and revisiting delayed consequences.

The diagnostics reveal three recurring failure patterns:
\begin{itemize}[leftmargin=*]
    \item \textbf{Incomplete evidence acquisition.} Agents often act on incomplete evidence, missing critical information for orders or price updates even when the corresponding tools are available.
    \item \textbf{Weak evidence-to-action conversion.} Agents rely on surface-level signals such as supplier price, while underweighting less visible but consequential factors such as supplier quality, product returns, expiration risk, and long-term profitability.
    \item \textbf{Lack of a consistent long-horizon policy.} Agents fail to maintain attention over important products and often do not follow up on delayed consequences of prior actions.
\end{itemize}
Detailed metric definitions are provided in Appendix~\ref{appendix:stage_metrics}.

\begin{figure*}[t]
    \centering

    \begin{minipage}{0.44\textwidth}
        \centering
        \includegraphics[width=0.97\linewidth]{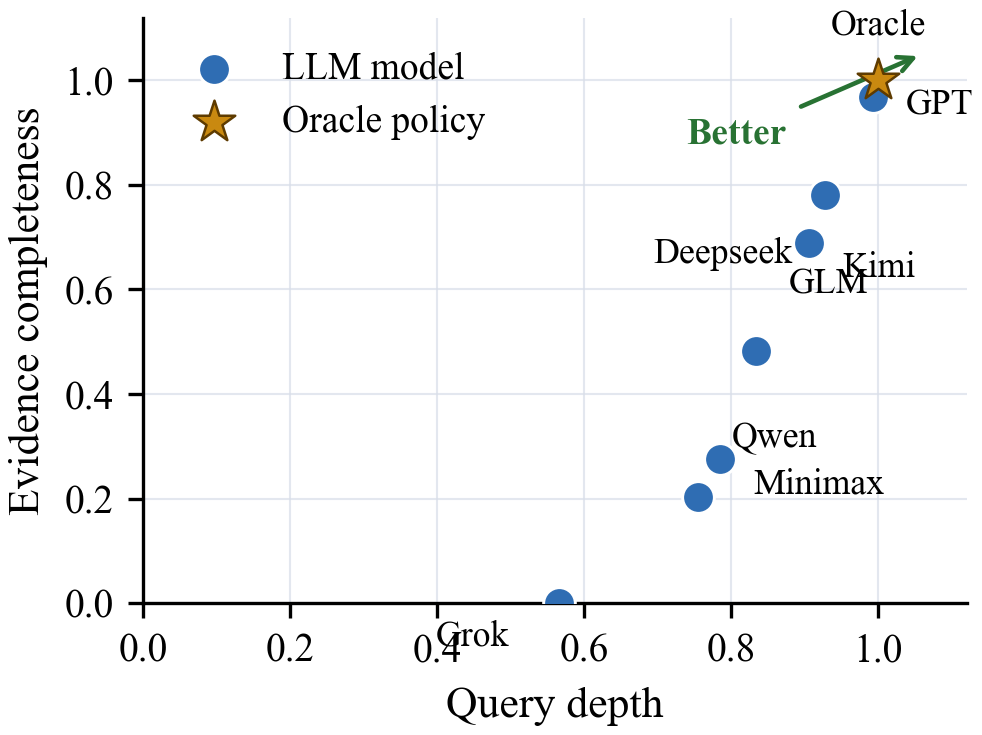}
        \vspace{0.1em}
        \centerline{\small \textbf{(a) Evidence acquisition}}
    \end{minipage}
    \hfill
    \begin{minipage}{0.44\textwidth}
        \centering
        \includegraphics[width=0.97\linewidth]{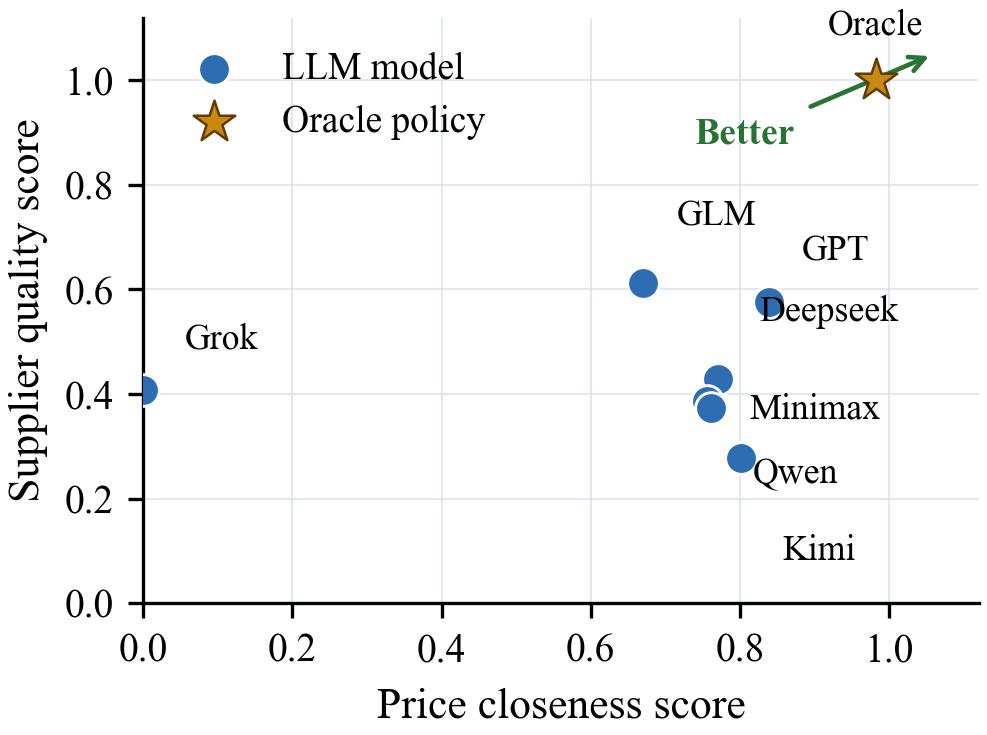}
        \vspace{0.1em}
        \centerline{\small \textbf{(b) Action conversion}}
    \end{minipage}

    \vspace{1.15em}

    \begin{minipage}{0.44\textwidth}
        \centering
        \includegraphics[width=0.97\linewidth]{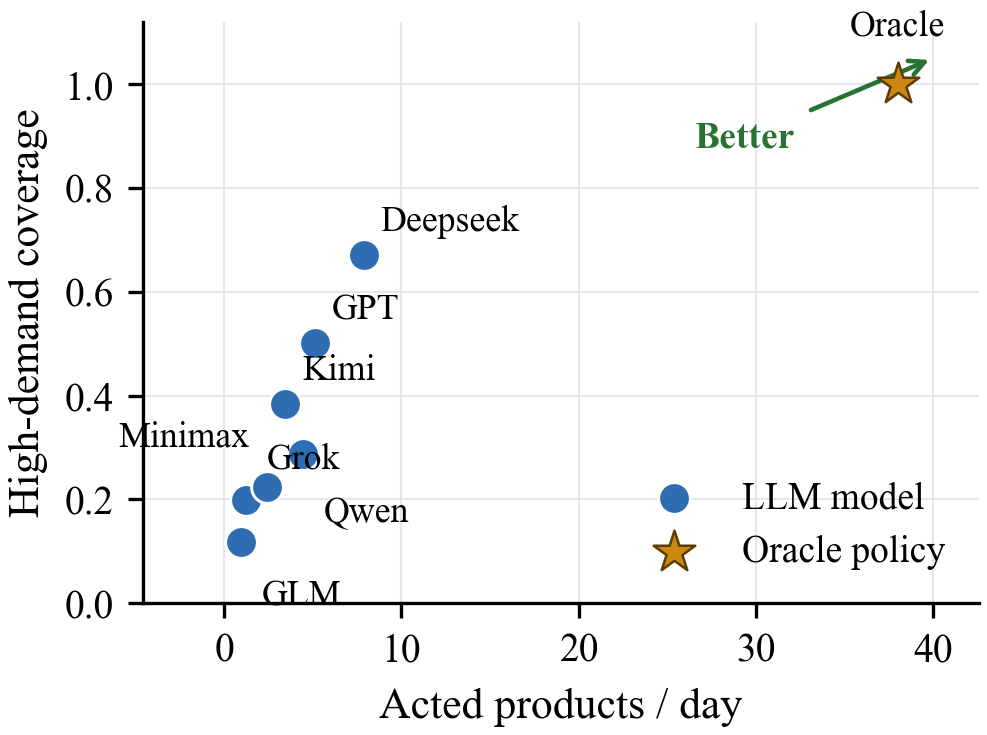}
        \vspace{0.1em}
        \centerline{\small \textbf{(c) Product candidate selection}}
    \end{minipage}
    \hfill
    \begin{minipage}{0.44\textwidth}
        \centering
        \includegraphics[width=0.97\linewidth]{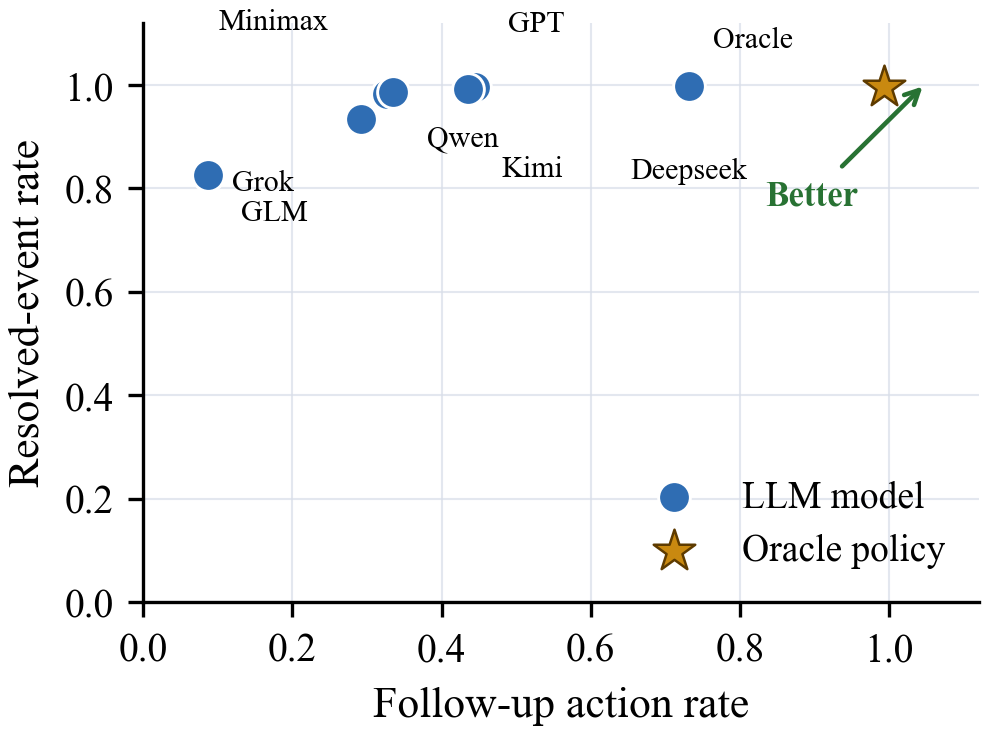}
        \vspace{0.1em}
        \centerline{\small \textbf{(d) Temporal follow-up}}
    \end{minipage}

    \caption{Diagnostic analysis of the three failure patterns in RetailBench. Blue circles denote survival-first selected LLM runs, and the black star denotes the oracle policy. Better performance appears toward the upper-right region in each panel. Panels (a) and (b) diagnose incomplete evidence acquisition and surface-level decision making, respectively; panels (c) and (d) diagnose the lack of a consistent long-horizon policy.}
    \label{fig:failure_diagnostics}
\end{figure*}

\subsection{Incomplete Evidence Acquisition}

The first failure pattern is incomplete evidence acquisition. Agents often place orders or update prices without inspecting key information about demand, inventory, supplier conditions, reviews, and external events. We use two diagnostics:
\begin{itemize}[leftmargin=*]
    \item \textbf{Query depth} ($\uparrow$): the fraction of required evidence categories queried before an order or price update.
    \item \textbf{Evidence completeness} ($\uparrow$): the fraction of actions for which all decision-critical evidence categories have been queried before execution.
\end{itemize}

Figure~\ref{fig:failure_diagnostics}(a) shows large variation in evidence acquisition. GPT-5.5 performs best, with a query depth of 0.9922 and an evidence completeness of 0.9689. DeepSeek-V4-Pro and Kimi-K2.6 also achieve high query depth, at 0.9279 and 0.9052, respectively. In contrast, MiniMax-M2.5, Qwen3.5, and Grok-4.3 show lower evidence completeness, with Grok-4.3 reaching 0.0000. This pattern broadly aligns with Table~\ref{tab:selected_main_results}, where agents with stronger evidence acquisition generally obtain better long-horizon outcomes. The result suggests that weakly structured evidence acquisition limits downstream action quality, though evidence collection alone does not guarantee effective decisions.

\subsection{Surface-Level Decision Making}

The second failure pattern is surface-level decision making. Agents often rely on immediately visible signals rather than criteria that determine long-term outcomes. In RetailBench, supplier price is directly observable, whereas supplier quality is only indirectly reflected through return rates and customer reviews. As a result, agents may favor low-cost suppliers even when lower quality can increase returns, weaken customer feedback, and reduce long-term profitability.

We use two diagnostics:
\begin{itemize}[leftmargin=*]
    \item \textbf{Price closeness score} ($\uparrow$): a normalized score derived from the distance between the chosen price and an estimated historical-profit optimum.
    \item \textbf{Supplier quality score} ($\uparrow$): a normalized rank-based score of the selected supplier among same-product candidates.
\end{itemize}

Figure~\ref{fig:failure_diagnostics}(b) shows that action conversion is the clearest bottleneck. On the price-closeness axis, the oracle policy corresponds to a mean raw price distance of 1.98\%, while the closest selected LLM run, GPT-5.5, still has a raw distance of 17.92\%. Supplier-quality scores remain consistently weak: in raw rank units, selected LLM mean quality ranks range from 2.55 to 3.89, and the best selected-run \textbf{QualityFirst} rate is only 34.65\%. Here, \textbf{QualityFirst} denotes choosing the raw-quality-best supplier among same-product candidates, whereas \textbf{PriceFirst} denotes choosing the cheapest supplier; Appendix~\ref{appendix:stage_metrics} and Figure~\ref{fig:supplier_selection_diagnostic} provide the detailed diagnostic definition and comparison. Across all LLM runs, \textbf{QualityFirst} is 21.5\%, compared with 55.6\% for \textbf{PriceFirst}.

Thus, the bottleneck is not merely tool use, but evidence-to-action reasoning. Agents often query supplier prices, yet less consistently query or use supplier-quality proxies, limiting their ability to make quality-adjusted decisions with better long-term outcomes.
\nolinenumbers
\subsection{Lack of Consistent Long-Horizon Policy}

The third failure pattern is the lack of a consistent stateful policy over long horizons. Long-horizon store operation is not a sequence of independent daily decisions: agents must maintain attention over important products, track prior actions, and update procurement, shelf, and pricing decisions after delayed feedback arrives. We diagnose this pattern from two complementary perspectives: product candidate selection and temporal follow-up.

For product candidate selection, we use two diagnostics:
\begin{itemize}[leftmargin=*]
    \item \textbf{Acted products per day} ($\uparrow$): the average number of products receiving information queries or management actions, such as orders or price updates, per day.
    \item \textbf{High-demand coverage} ($\uparrow$): the extent to which daily top-sales or stockout products receive attention.
\end{itemize}

Figure~\ref{fig:failure_diagnostics}(c) shows insufficient product-level attention across LLM agents. The oracle policy acts on 38.00 products per day and misses no retrospective high-demand opportunities. Selected LLM runs, however, act on only 0.95 to 7.89 products per day and miss 32.86\% to 88.18\% of high-demand opportunities. Even the strongest selected LLM on this diagnostic, DeepSeek-V4-Pro, acts on only 7.89 products per day and misses 32.86\% of high-demand opportunities.

These results indicate that many failures arise before action quality is considered: although daily sales may span more than 30 distinct products, most sold products simply do not enter the daily action set. Persistent narrowness in product attention reflects weak portfolio-level management, rather than a one-step action error.

For temporal follow-up, we use two diagnostics:
\begin{itemize}[leftmargin=*]
    \item \textbf{Follow-up action rate} ($\uparrow$): the fraction of acted products that receive another action within seven days.
    \item \textbf{Resolved-event rate} ($\uparrow$): one minus the rate of stockout, return, or expiration events left without later attention.
\end{itemize}

Figure~\ref{fig:failure_diagnostics}(d) shows substantial variation in follow-up quality. The oracle policy follows up on 0.9936 of acted products within seven days and leaves only 0.0040 of delayed events unresolved. DeepSeek-V4-Pro has the strongest selected-LLM follow-up action rate, 0.7316, and an unresolved-event rate of 0.0028, consistent with its 180-day survival. GPT-5.5 also keeps unresolved events low at 0.0042, but its follow-up action rate is lower at 0.4443.

Together, product selection and temporal follow-up show that long-horizon failures are not merely poor one-step decisions. Agents lack persistent stateful policies that maintain product attention, revisit delayed consequences, and revise operational decisions after feedback arrives. This explains why some agents make occasional reasonable local decisions but still fail to sustain stable store operation over the full horizon.

Overall, systematic LLM failure in RetailBench arises from three interacting bottlenecks: incomplete evidence acquisition, weak evidence-to-action conversion, and inconsistent long-horizon policy. Improving long-horizon LLM agents therefore requires more than increased tool use; it also requires better evidence selection, deeper evidence-to-action reasoning, and more reliable stateful decision policies.
\FloatBarrier

\section{Related Work}
\label{sec:related_work}
\paragraph{Retail, e-commerce, and supply-chain decision benchmarks.}
Prior work has studied retail, e-commerce, and supply-chain decision-making through benchmarks for retail simulation and promotion optimization~\citep{xia2023retailsynth,xia2024simulationretailpromotions}, demand recovery and transaction modeling~\citep{wang2025freshretailnet,tkachuk2024consumertransactions}, web-shopping agents~\citep{wang2025shoppingbench,3webshopscalablerealworldweb}, and supply-chain or inventory control~\citep{OFCOURSE,inventory,leluc2023marlimmultiagentreinforcementlearning,invagentlargelanguagemodel,aimbenchevaluatingdecisionmakingbiases}. These benchmarks cover important subproblems but typically isolate promotion, demand modeling, shopping, or inventory decisions. RetailBench instead evaluates tool-using LLM agents in a coupled retail environment where pricing, replenishment, supplier choice, shelf assortment, feedback, events, and cash flow interact over time.

\paragraph{Long-horizon agent benchmarks.}
Recent benchmarks increasingly evaluate LLM agents beyond isolated task completion, including ultra-long-horizon interaction, office workflows, virtual-world planning, lifelong experience reuse, workplace task execution, and long-running business operation~\citep{utltabench,odysseybenchevaluatingllmagents,herobenchbenchmarklonghorizonplanning,zheng2025lifelongagentbench,xu2024theagentcompany,backlund2025vendingbenchbenchmarklongtermcoherence,andonlabs2025vendingbench2}. They highlight persistence, memory, exploration, and workflow-level decision making, but generally do not model data-grounded retail dynamics over many days. RetailBench complements this line by grounding evaluation in closed-loop product-level operation and supporting diagnostic analysis of evidence acquisition, evidence-to-action conversion, and long-horizon policy consistency.

\section{Conclusion}
\label{sec:conclusion}
We introduced \textbf{RetailBench}, a data-grounded benchmark for evaluating long-horizon, tool-using LLM agents in closed-loop retail operation. RetailBench couples pricing, replenishment, supplier selection, shelf assortment, customer feedback, external events, and financial constraints over extended episodes. Our results show that current LLM agents still struggle to sustain effective operation: even models that survive the full horizon remain far below the oracle policy in net worth and sales outcomes. Failure analysis identifies three recurring bottlenecks: incomplete evidence acquisition, surface-level decision making, and lack of a consistent long-horizon policy. RetailBench therefore serves not only as a performance benchmark, but also as a diagnostic testbed for whether agents can maintain operational state, revisit delayed consequences, and align local tool calls with long-term business outcomes. These findings suggest that future agents need stronger evidence selection, evidence-to-action reasoning, stateful policy maintenance, and feedback adaptation before they can reliably operate in complex long-horizon environments.

\noindent\textbf{Code and data availability.}
Code, benchmark materials, processed data, and reproduction instructions are
available at \url{https://github.com/linghuazhang01/RetailBench}.

\section{Limitations}
\label{sec:limitation}
RetailBench has several limitations. It focuses on a single-store supermarket setting and therefore omits multi-store coordination, competitive markets, strategic supplier behavior, labor constraints, promotions, and interactions among multiple autonomous agents. Although the environment is grounded in real product and sales signals, reviews, news events, supplier adaptation, returns, and parts of the demand model are synthetic or rule-based, so RetailBench should be viewed as a controlled stress test rather than a full-fidelity simulator of retail economics. Our evaluation is also limited to prompting-based agents under a fixed environment configuration, and the selected-run protocol characterizes the strongest observed behavior rather than average performance across all possible agent designs or environment settings. Finally, the failure analysis is diagnostic rather than causal: the proposed metrics identify recurring bottlenecks but do not fully isolate each factor's causal contribution. Future work should extend RetailBench to richer retail settings, multi-agent operation, learning-based adaptation, counterfactual diagnostics, and constraint-aware action control.

\bibliography{custom}

@misc{DominicksDataset,
  author = {{Kilts Center for Marketing, University of Chicago Booth School of Business}},
  title = {{Dominick's Dataset}},
  year = {n.d.},
  howpublished = {\url{https://www.chicagobooth.edu/research/kilts/research-data/dominicks}},
  note = {Accessed: 2025-11-01}
}

@incollection{mcfadden_conditional_1974,
	address = {New York},
	title = {Conditional logit analysis of qualitative choice behavior},
	language = {en},
	booktitle = {Frontiers in {Econometrics}},
	publisher = {Academic press},
	author = {McFadden, Daniel},
	editor = {Zarembka, Paul},
	year = {1974},
	pages = {105--142}
}

@misc{hou2024bridging,
  title = {Bridging Language and Items for Retrieval and Recommendation},
  author = {Hou, Yupeng and Li, Jiacheng and He, Zhankui and Yan, An and Chen, Xiusi and McAuley, Julian},
  year = {2024},
  eprint = {2403.03952},
  archivePrefix = {arXiv},
  primaryClass = {cs.IR},
  url = {https://arxiv.org/abs/2403.03952}
}

@misc{ashraq2025financialNewsArticles,
  title        = {financial-news-articles},
  author       = {ashraq},
  year         = {2025},
  howpublished = {\url{https://huggingface.co/datasets/ashraq/financial-news-articles}},
  note         = {Accessed: 2025-12-01}
}

@article{nakanishi1974mci,
  author = {Nakanishi, Masao and Cooper, Lee G.},
  title = {Parameter Estimation for a Multiplicative Competitive Interaction Model---Least Squares Approach},
  journal = {Journal of Marketing Research},
  volume = {11},
  number = {3},
  pages = {303--311},
  year = {1974},
  doi = {10.1177/002224377401100309}
}

@misc{amodei2024machines,
  author       = {Dario Amodei},
  title        = {Machines of Loving Grace},
  year         = {2024},
  howpublished = {\url{https://www.darioamodei.com/essay/machines-of-loving-grace}},
  note         = {Accessed: 2025-12-01}
}

@misc{backlund2025vendingbenchbenchmarklongtermcoherence,
      title={Vending-Bench: A Benchmark for Long-Term Coherence of Autonomous Agents}, 
      author={Axel Backlund and Lukas Petersson},
      year={2025},
      eprint={2502.15840},
      archivePrefix={arXiv},
      primaryClass={cs.AI},
      url={https://arxiv.org/abs/2502.15840}, 
}

@misc{andonlabs2025vendingbench2,
  title        = {Vending-Bench 2: A Benchmark for Long-Horizon Business Simulation},
  author       = {{Andon Labs}},
  year         = {2025},
  howpublished = {\url{https://andonlabs.com/evals/vending-bench-2}},
  note         = {Accessed: 2025-12-10}
}

@misc{kwa2025measuringaiabilitycomplete,
      title={Measuring AI Ability to Complete Long Tasks}, 
      author={Thomas Kwa and Ben West and Joel Becker and Amy Deng and Katharyn Garcia and Max Hasin and Sami Jawhar and Megan Kinniment and Nate Rush and Sydney Von Arx and Ryan Bloom and Thomas Broadley and Haoxing Du and Brian Goodrich and Nikola Jurkovic and Luke Harold Miles and Seraphina Nix and Tao Lin and Neev Parikh and David Rein and Lucas Jun Koba Sato and Hjalmar Wijk and Daniel M. Ziegler and Elizabeth Barnes and Lawrence Chan},
      year={2025},
      eprint={2503.14499},
      archivePrefix={arXiv},
      primaryClass={cs.AI},
      url={https://arxiv.org/abs/2503.14499}, 
}

@inproceedings{swe-bench,
  title = {{SWE-bench}: Can Language Models Resolve Real-World {GitHub} Issues?},
  author = {Jimenez, Carlos E. and Yang, John and Wettig, Alexander and Yao, Shunyu and Pei, Kexin and Press, Ofir and Narasimhan, Karthik},
  booktitle = {International Conference on Learning Representations},
  year = {2024},
  eprint = {2310.06770},
  archivePrefix = {arXiv},
  primaryClass = {cs.CL},
  url = {https://proceedings.iclr.cc/paper_files/paper/2024/hash/edac78c3e300629acfe6cbe9ca88fb84-Abstract-Conference.html}
}

@misc{hle,
      title={Humanity's Last Exam}, 
      author={Long Phan and Alice Gatti and Ziwen Han and Nathaniel Li and Josephina Hu and Hugh Zhang and Chen Bo Calvin Zhang and Mohamed Shaaban and John Ling and Sean Shi and Michael Choi and Anish Agrawal and Arnav Chopra and Adam Khoja and Ryan Kim and Richard Ren and Jason Hausenloy and Oliver Zhang and Mantas Mazeika and Dmitry Dodonov and Tung Nguyen and Jaeho Lee and Daron Anderson and Mikhail Doroshenko and Alun Cennyth Stokes and Mobeen Mahmood and Oleksandr Pokutnyi and Oleg Iskra and Jessica P. Wang and John-Clark Levin and Mstyslav Kazakov and Fiona Feng and Steven Y. Feng and Haoran Zhao and Michael Yu and Varun Gangal and Chelsea Zou and Zihan Wang and Serguei Popov and Robert Gerbicz and Geoff Galgon and Johannes Schmitt and Will Yeadon and Yongki Lee and Scott Sauers and Alvaro Sanchez and Fabian Giska and Marc Roth and Søren Riis and Saiteja Utpala and Noah Burns and Gashaw M. Goshu and Mohinder Maheshbhai Naiya and Chidozie Agu and Zachary Giboney and Antrell Cheatom and Francesco Fournier-Facio and Sarah-Jane Crowson and Lennart Finke and Zerui Cheng and Jennifer Zampese and Ryan G. Hoerr and Mark Nandor and Hyunwoo Park and Tim Gehrunger and Jiaqi Cai and Ben McCarty and Alexis C Garretson and Edwin Taylor and Damien Sileo and Qiuyu Ren and Usman Qazi and Lianghui Li and Jungbae Nam and John B. Wydallis and Pavel Arkhipov and Jack Wei Lun Shi and Aras Bacho and Chris G. Willcocks and Hangrui Cao and Sumeet Motwani and Emily de Oliveira Santos and Johannes Veith and Edward Vendrow and Doru Cojoc and Kengo Zenitani and Joshua Robinson and Longke Tang and Yuqi Li and Joshua Vendrow and Natanael Wildner Fraga and Vladyslav Kuchkin and Andrey Pupasov Maksimov and Pierre Marion and Denis Efremov and Jayson Lynch and Kaiqu Liang and Aleksandar Mikov and Andrew Gritsevskiy and Julien Guillod and Gözdenur Demir and Dakotah Martinez and Ben Pageler and Kevin Zhou and Saeed Soori and Ori Press and Henry Tang and Paolo Rissone and Sean R. Green and Lina Brüssel and Moon Twayana and Aymeric Dieuleveut and Joseph Marvin Imperial and Ameya Prabhu and Jinzhou Yang and Nick Crispino and Arun Rao and Dimitri Zvonkine and Gabriel Loiseau and Mikhail Kalinin and Marco Lukas and Ciprian Manolescu and Nate Stambaugh and Subrata Mishra and Tad Hogg and Carlo Bosio and Brian P Coppola and Julian Salazar and Jaehyeok Jin and Rafael Sayous and Stefan Ivanov and Philippe Schwaller and Shaipranesh Senthilkuma and Andres M Bran and Andres Algaba and Kelsey Van den Houte and Lynn Van Der Sypt and Brecht Verbeken and David Noever and Alexei Kopylov and Benjamin Myklebust and Bikun Li and Lisa Schut and Evgenii Zheltonozhskii and Qiaochu Yuan and Derek Lim and Richard Stanley and Tong Yang and John Maar and Julian Wykowski and Martí Oller and Anmol Sahu and Cesare Giulio Ardito and Yuzheng Hu and Ariel Ghislain Kemogne Kamdoum and Alvin Jin and Tobias Garcia Vilchis and Yuexuan Zu and Martin Lackner and James Koppel and Gongbo Sun and Daniil S. Antonenko and Steffi Chern and Bingchen Zhao and Pierrot Arsene and Joseph M Cavanagh and Daofeng Li and Jiawei Shen and Donato Crisostomi and Wenjin Zhang and Ali Dehghan and Sergey Ivanov and David Perrella and Nurdin Kaparov and Allen Zang and Ilia Sucholutsky and Arina Kharlamova and Daniil Orel and Vladislav Poritski and Shalev Ben-David and Zachary Berger and Parker Whitfill and Michael Foster and Daniel Munro and Linh Ho and Shankar Sivarajan and Dan Bar Hava and Aleksey Kuchkin and David Holmes and Alexandra Rodriguez-Romero and Frank Sommerhage and Anji Zhang and Richard Moat and Keith Schneider and Zakayo Kazibwe and Don Clarke and Dae Hyun Kim and Felipe Meneguitti Dias and Sara Fish and Veit Elser and Tobias Kreiman and Victor Efren Guadarrama Vilchis and Immo Klose and Ujjwala Anantheswaran and Adam Zweiger and Kaivalya Rawal and Jeffery Li and Jeremy Nguyen and Nicolas Daans and Haline Heidinger and Maksim Radionov and Václav Rozhoň and Vincent Ginis and Christian Stump and Niv Cohen and Rafał Poświata and Josef Tkadlec and Alan Goldfarb and Chenguang Wang and Piotr Padlewski and Stanislaw Barzowski and Kyle Montgomery and Ryan Stendall and Jamie Tucker-Foltz and Jack Stade and T. Ryan Rogers and Tom Goertzen and Declan Grabb and Abhishek Shukla and Alan Givré and John Arnold Ambay and Archan Sen and Muhammad Fayez Aziz and Mark H Inlow and Hao He and Ling Zhang and Younesse Kaddar and Ivar Ängquist and Yanxu Chen and Harrison K Wang and Kalyan Ramakrishnan and Elliott Thornley and Antonio Terpin and Hailey Schoelkopf and Eric Zheng and Avishy Carmi and Ethan D. L. Brown and Kelin Zhu and Max Bartolo and Richard Wheeler and Martin Stehberger and Peter Bradshaw and JP Heimonen and Kaustubh Sridhar and Ido Akov and Jennifer Sandlin and Yury Makarychev and Joanna Tam and Hieu Hoang and David M. Cunningham and Vladimir Goryachev and Demosthenes Patramanis and Michael Krause and Andrew Redenti and David Aldous and Jesyin Lai and Shannon Coleman and Jiangnan Xu and Sangwon Lee and Ilias Magoulas and Sandy Zhao and Ning Tang and Michael K. Cohen and Orr Paradise and Jan Hendrik Kirchner and Maksym Ovchynnikov and Jason O. Matos and Adithya Shenoy and Michael Wang and Yuzhou Nie and Anna Sztyber-Betley and Paolo Faraboschi and Robin Riblet and Jonathan Crozier and Shiv Halasyamani and Shreyas Verma and Prashant Joshi and Eli Meril and Ziqiao Ma and Jérémy Andréoletti and Raghav Singhal and Jacob Platnick and Volodymyr Nevirkovets and Luke Basler and Alexander Ivanov and Seri Khoury and Nils Gustafsson and Marco Piccardo and Hamid Mostaghimi and Qijia Chen and Virendra Singh and Tran Quoc Khánh and Paul Rosu and Hannah Szlyk and Zachary Brown and Himanshu Narayan and Aline Menezes and Jonathan Roberts and William Alley and Kunyang Sun and Arkil Patel and Max Lamparth and Anka Reuel and Linwei Xin and Hanmeng Xu and Jacob Loader and Freddie Martin and Zixuan Wang and Andrea Achilleos and Thomas Preu and Tomek Korbak and Ida Bosio and Fereshteh Kazemi and Ziye Chen and Biró Bálint and Eve J. Y. Lo and Jiaqi Wang and Maria Inês S. Nunes and Jeremiah Milbauer and M Saiful Bari and Zihao Wang and Behzad Ansarinejad and Yewen Sun and Stephane Durand and Hossam Elgnainy and Guillaume Douville and Daniel Tordera and George Balabanian and Hew Wolff and Lynna Kvistad and Hsiaoyun Milliron and Ahmad Sakor and Murat Eron and Andrew Favre D. O. and Shailesh Shah and Xiaoxiang Zhou and Firuz Kamalov and Sherwin Abdoli and Tim Santens and Shaul Barkan and Allison Tee and Robin Zhang and Alessandro Tomasiello and G. Bruno De Luca and Shi-Zhuo Looi and Vinh-Kha Le and Noam Kolt and Jiayi Pan and Emma Rodman and Jacob Drori and Carl J Fossum and Niklas Muennighoff and Milind Jagota and Ronak Pradeep and Honglu Fan and Jonathan Eicher and Michael Chen and Kushal Thaman and William Merrill and Moritz Firsching and Carter Harris and Stefan Ciobâcă and Jason Gross and Rohan Pandey and Ilya Gusev and Adam Jones and Shashank Agnihotri and Pavel Zhelnov and Mohammadreza Mofayezi and Alexander Piperski and David K. Zhang and Kostiantyn Dobarskyi and Roman Leventov and Ignat Soroko and Joshua Duersch and Vage Taamazyan and Andrew Ho and Wenjie Ma and William Held and Ruicheng Xian and Armel Randy Zebaze and Mohanad Mohamed and Julian Noah Leser and Michelle X Yuan and Laila Yacar and Johannes Lengler and Katarzyna Olszewska and Claudio Di Fratta and Edson Oliveira and Joseph W. Jackson and Andy Zou and Muthu Chidambaram and Timothy Manik and Hector Haffenden and Dashiell Stander and Ali Dasouqi and Alexander Shen and Bita Golshani and David Stap and Egor Kretov and Mikalai Uzhou and Alina Borisovna Zhidkovskaya and Nick Winter and Miguel Orbegozo Rodriguez and Robert Lauff and Dustin Wehr and Colin Tang and Zaki Hossain and Shaun Phillips and Fortuna Samuele and Fredrik Ekström and Angela Hammon and Oam Patel and Faraz Farhidi and George Medley and Forough Mohammadzadeh and Madellene Peñaflor and Haile Kassahun and Alena Friedrich and Rayner Hernandez Perez and Daniel Pyda and Taom Sakal and Omkar Dhamane and Ali Khajegili Mirabadi and Eric Hallman and Kenchi Okutsu and Mike Battaglia and Mohammad Maghsoudimehrabani and Alon Amit and Dave Hulbert and Roberto Pereira and Simon Weber and Handoko and Anton Peristyy and Stephen Malina and Mustafa Mehkary and Rami Aly and Frank Reidegeld and Anna-Katharina Dick and Cary Friday and Mukhwinder Singh and Hassan Shapourian and Wanyoung Kim and Mariana Costa and Hubeyb Gurdogan and Harsh Kumar and Chiara Ceconello and Chao Zhuang and Haon Park and Micah Carroll and Andrew R. Tawfeek and Stefan Steinerberger and Daattavya Aggarwal and Michael Kirchhof and Linjie Dai and Evan Kim and Johan Ferret and Jainam Shah and Yuzhou Wang and Minghao Yan and Krzysztof Burdzy and Lixin Zhang and Antonio Franca and Diana T. Pham and Kang Yong Loh and Joshua Robinson and Abram Jackson and Paolo Giordano and Philipp Petersen and Adrian Cosma and Jesus Colino and Colin White and Jacob Votava and Vladimir Vinnikov and Ethan Delaney and Petr Spelda and Vit Stritecky and Syed M. Shahid and Jean-Christophe Mourrat and Lavr Vetoshkin and Koen Sponselee and Renas Bacho and Zheng-Xin Yong and Florencia de la Rosa and Nathan Cho and Xiuyu Li and Guillaume Malod and Orion Weller and Guglielmo Albani and Leon Lang and Julien Laurendeau and Dmitry Kazakov and Fatimah Adesanya and Julien Portier and Lawrence Hollom and Victor Souza and Yuchen Anna Zhou and Julien Degorre and Yiğit Yalın and Gbenga Daniel Obikoya and Rai and Filippo Bigi and M. C. Boscá and Oleg Shumar and Kaniuar Bacho and Gabriel Recchia and Mara Popescu and Nikita Shulga and Ngefor Mildred Tanwie and Thomas C. H. Lux and Ben Rank and Colin Ni and Matthew Brooks and Alesia Yakimchyk and Huanxu and Liu and Stefano Cavalleri and Olle Häggström and Emil Verkama and Joshua Newbould and Hans Gundlach and Leonor Brito-Santana and Brian Amaro and Vivek Vajipey and Rynaa Grover and Ting Wang and Yosi Kratish and Wen-Ding Li and Sivakanth Gopi and Andrea Caciolai and Christian Schroeder de Witt and Pablo Hernández-Cámara and Emanuele Rodolà and Jules Robins and Dominic Williamson and Vincent Cheng and Brad Raynor and Hao Qi and Ben Segev and Jingxuan Fan and Sarah Martinson and Erik Y. Wang and Kaylie Hausknecht and Michael P. Brenner and Mao Mao and Christoph Demian and Peyman Kassani and Xinyu Zhang and David Avagian and Eshawn Jessica Scipio and Alon Ragoler and Justin Tan and Blake Sims and Rebeka Plecnik and Aaron Kirtland and Omer Faruk Bodur and D. P. Shinde and Yan Carlos Leyva Labrador and Zahra Adoul and Mohamed Zekry and Ali Karakoc and Tania C. B. Santos and Samir Shamseldeen and Loukmane Karim and Anna Liakhovitskaia and Nate Resman and Nicholas Farina and Juan Carlos Gonzalez and Gabe Maayan and Earth Anderson and Rodrigo De Oliveira Pena and Elizabeth Kelley and Hodjat Mariji and Rasoul Pouriamanesh and Wentao Wu and Ross Finocchio and Ismail Alarab and Joshua Cole and Danyelle Ferreira and Bryan Johnson and Mohammad Safdari and Liangti Dai and Siriphan Arthornthurasuk and Isaac C. McAlister and Alejandro José Moyano and Alexey Pronin and Jing Fan and Angel Ramirez-Trinidad and Yana Malysheva and Daphiny Pottmaier and Omid Taheri and Stanley Stepanic and Samuel Perry and Luke Askew and Raúl Adrián Huerta Rodríguez and Ali M. R. Minissi and Ricardo Lorena and Krishnamurthy Iyer and Arshad Anil Fasiludeen and Ronald Clark and Josh Ducey and Matheus Piza and Maja Somrak and Eric Vergo and Juehang Qin and Benjámin Borbás and Eric Chu and Jack Lindsey and Antoine Jallon and I. M. J. McInnis and Evan Chen and Avi Semler and Luk Gloor and Tej Shah and Marc Carauleanu and Pascal Lauer and Tran Đuc Huy and Hossein Shahrtash and Emilien Duc and Lukas Lewark and Assaf Brown and Samuel Albanie and Brian Weber and Warren S. Vaz and Pierre Clavier and Yiyang Fan and Gabriel Poesia Reis e Silva and Long and Lian and Marcus Abramovitch and Xi Jiang and Sandra Mendoza and Murat Islam and Juan Gonzalez and Vasilios Mavroudis and Justin Xu and Pawan Kumar and Laxman Prasad Goswami and Daniel Bugas and Nasser Heydari and Ferenc Jeanplong and Thorben Jansen and Antonella Pinto and Archimedes Apronti and Abdallah Galal and Ng Ze-An and Ankit Singh and Tong Jiang and Joan of Arc Xavier and Kanu Priya Agarwal and Mohammed Berkani and Gang Zhang and Zhehang Du and Benedito Alves de Oliveira Junior and Dmitry Malishev and Nicolas Remy and Taylor D. Hartman and Tim Tarver and Stephen Mensah and Gautier Abou Loume and Wiktor Morak and Farzad Habibi and Sarah Hoback and Will Cai and Javier Gimenez and Roselynn Grace Montecillo and Jakub Łucki and Russell Campbell and Asankhaya Sharma and Khalida Meer and Shreen Gul and Daniel Espinosa Gonzalez and Xavier Alapont and Alex Hoover and Gunjan Chhablani and Freddie Vargus and Arunim Agarwal and Yibo Jiang and Deepakkumar Patil and David Outevsky and Kevin Joseph Scaria and Rajat Maheshwari and Abdelkader Dendane and Priti Shukla and Ashley Cartwright and Sergei Bogdanov and Niels Mündler and Sören Möller and Luca Arnaboldi and Kunvar Thaman and Muhammad Rehan Siddiqi and Prajvi Saxena and Himanshu Gupta and Tony Fruhauff and Glen Sherman and Mátyás Vincze and Siranut Usawasutsakorn and Dylan Ler and Anil Radhakrishnan and Innocent Enyekwe and Sk Md Salauddin and Jiang Muzhen and Aleksandr Maksapetyan and Vivien Rossbach and Chris Harjadi and Mohsen Bahaloohoreh and Claire Sparrow and Jasdeep Sidhu and Sam Ali and Song Bian and John Lai and Eric Singer and Justine Leon Uro and Greg Bateman and Mohamed Sayed and Ahmed Menshawy and Darling Duclosel and Dario Bezzi and Yashaswini Jain and Ashley Aaron and Murat Tiryakioglu and Sheeshram Siddh and Keith Krenek and Imad Ali Shah and Jun Jin and Scott Creighton and Denis Peskoff and Zienab EL-Wasif and Ragavendran P V and Michael Richmond and Joseph McGowan and Tejal Patwardhan and Hao-Yu Sun and Ting Sun and Nikola Zubić and Samuele Sala and Stephen Ebert and Jean Kaddour and Manuel Schottdorf and Dianzhuo Wang and Gerol Petruzella and Alex Meiburg and Tilen Medved and Ali ElSheikh and S Ashwin Hebbar and Lorenzo Vaquero and Xianjun Yang and Jason Poulos and Vilém Zouhar and Sergey Bogdanik and Mingfang Zhang and Jorge Sanz-Ros and David Anugraha and Yinwei Dai and Anh N. Nhu and Xue Wang and Ali Anil Demircali and Zhibai Jia and Yuyin Zhou and Juncheng Wu and Mike He and Nitin Chandok and Aarush Sinha and Gaoxiang Luo and Long Le and Mickaël Noyé and Michał Perełkiewicz and Ioannis Pantidis and Tianbo Qi and Soham Sachin Purohit and Letitia Parcalabescu and Thai-Hoa Nguyen and Genta Indra Winata and Edoardo M. Ponti and Hanchen Li and Kaustubh Dhole and Jongee Park and Dario Abbondanza and Yuanli Wang and Anupam Nayak and Diogo M. Caetano and Antonio A. W. L. Wong and Maria del Rio-Chanona and Dániel Kondor and Pieter Francois and Ed Chalstrey and Jakob Zsambok and Dan Hoyer and Jenny Reddish and Jakob Hauser and Francisco-Javier Rodrigo-Ginés and Suchandra Datta and Maxwell Shepherd and Thom Kamphuis and Qizheng Zhang and Hyunjun Kim and Ruiji Sun and Jianzhu Yao and Franck Dernoncourt and Satyapriya Krishna and Sina Rismanchian and Bonan Pu and Francesco Pinto and Yingheng Wang and Kumar Shridhar and Kalon J. Overholt and Glib Briia and Hieu Nguyen and David and Soler Bartomeu and Tony CY Pang and Adam Wecker and Yifan Xiong and Fanfei Li and Lukas S. Huber and Joshua Jaeger and Romano De Maddalena and Xing Han Lù and Yuhui Zhang and Claas Beger and Patrick Tser Jern Kon and Sean Li and Vivek Sanker and Ming Yin and Yihao Liang and Xinlu Zhang and Ankit Agrawal and Li S. Yifei and Zechen Zhang and Mu Cai and Yasin Sonmez and Costin Cozianu and Changhao Li and Alex Slen and Shoubin Yu and Hyun Kyu Park and Gabriele Sarti and Marcin Briański and Alessandro Stolfo and Truong An Nguyen and Mike Zhang and Yotam Perlitz and Jose Hernandez-Orallo and Runjia Li and Amin Shabani and Felix Juefei-Xu and Shikhar Dhingra and Orr Zohar and My Chiffon Nguyen and Alexander Pondaven and Abdurrahim Yilmaz and Xuandong Zhao and Chuanyang Jin and Muyan Jiang and Stefan Todoran and Xinyao Han and Jules Kreuer and Brian Rabern and Anna Plassart and Martino Maggetti and Luther Yap and Robert Geirhos and Jonathon Kean and Dingsu Wang and Sina Mollaei and Chenkai Sun and Yifan Yin and Shiqi Wang and Rui Li and Yaowen Chang and Anjiang Wei and Alice Bizeul and Xiaohan Wang and Alexandre Oliveira Arrais and Kushin Mukherjee and Jorge Chamorro-Padial and Jiachen Liu and Xingyu Qu and Junyi Guan and Adam Bouyamourn and Shuyu Wu and Martyna Plomecka and Junda Chen and Mengze Tang and Jiaqi Deng and Shreyas Subramanian and Haocheng Xi and Haoxuan Chen and Weizhi Zhang and Yinuo Ren and Haoqin Tu and Sejong Kim and Yushun Chen and Sara Vera Marjanović and Junwoo Ha and Grzegorz Luczyna and Jeff J. Ma and Zewen Shen and Dawn Song and Cedegao E. Zhang and Zhun Wang and Gaël Gendron and Yunze Xiao and Leo Smucker and Erica Weng and Kwok Hao Lee and Zhe Ye and Stefano Ermon and Ignacio D. Lopez-Miguel and Theo Knights and Anthony Gitter and Namkyu Park and Boyi Wei and Hongzheng Chen and Kunal Pai and Ahmed Elkhanany and Han Lin and Philipp D. Siedler and Jichao Fang and Ritwik Mishra and Károly Zsolnai-Fehér and Xilin Jiang and Shadab Khan and Jun Yuan and Rishab Kumar Jain and Xi Lin and Mike Peterson and Zhe Wang and Aditya Malusare and Maosen Tang and Isha Gupta and Ivan Fosin and Timothy Kang and Barbara Dworakowska and Kazuki Matsumoto and Guangyao Zheng and Gerben Sewuster and Jorge Pretel Villanueva and Ivan Rannev and Igor Chernyavsky and Jiale Chen and Deepayan Banik and Ben Racz and Wenchao Dong and Jianxin Wang and Laila Bashmal and Duarte V. Gonçalves and Wei Hu and Kaushik Bar and Ondrej Bohdal and Atharv Singh Patlan and Shehzaad Dhuliawala and Caroline Geirhos and Julien Wist and Yuval Kansal and Bingsen Chen and Kutay Tire and Atak Talay Yücel and Brandon Christof and Veerupaksh Singla and Zijian Song and Sanxing Chen and Jiaxin Ge and Kaustubh Ponkshe and Isaac Park and Tianneng Shi and Martin Q. Ma and Joshua Mak and Sherwin Lai and Antoine Moulin and Zhuo Cheng and Zhanda Zhu and Ziyi Zhang and Vaidehi Patil and Ketan Jha and Qiutong Men and Jiaxuan Wu and Tianchi Zhang and Bruno Hebling Vieira and Alham Fikri Aji and Jae-Won Chung and Mohammed Mahfoud and Ha Thi Hoang and Marc Sperzel and Wei Hao and Kristof Meding and Sihan Xu and Vassilis Kostakos and Davide Manini and Yueying Liu and Christopher Toukmaji and Jay Paek and Eunmi Yu and Arif Engin Demircali and Zhiyi Sun and Ivan Dewerpe and Hongsen Qin and Roman Pflugfelder and James Bailey and Johnathan Morris and Ville Heilala and Sybille Rosset and Zishun Yu and Peter E. Chen and Woongyeong Yeo and Eeshaan Jain and Ryan Yang and Sreekar Chigurupati and Julia Chernyavsky and Sai Prajwal Reddy and Subhashini Venugopalan and Hunar Batra and Core Francisco Park and Hieu Tran and Guilherme Maximiano and Genghan Zhang and Yizhuo Liang and Hu Shiyu and Rongwu Xu and Rui Pan and Siddharth Suresh and Ziqi Liu and Samaksh Gulati and Songyang Zhang and Peter Turchin and Christopher W. Bartlett and Christopher R. Scotese and Phuong M. Cao and Ben Wu and Jacek Karwowski and Davide Scaramuzza and Aakaash Nattanmai and Gordon McKellips and Anish Cheraku and Asim Suhail and Ethan Luo and Marvin Deng and Jason Luo and Ashley Zhang and Kavin Jindel and Jay Paek and Kasper Halevy and Allen Baranov and Michael Liu and Advaith Avadhanam and David Zhang and Vincent Cheng and Brad Ma and Evan Fu and Liam Do and Joshua Lass and Hubert Yang and Surya Sunkari and Vishruth Bharath and Violet Ai and James Leung and Rishit Agrawal and Alan Zhou and Kevin Chen and Tejas Kalpathi and Ziqi Xu and Gavin Wang and Tyler Xiao and Erik Maung and Sam Lee and Ryan Yang and Roy Yue and Ben Zhao and Julia Yoon and Sunny Sun and Aryan Singh and Ethan Luo and Clark Peng and Tyler Osbey and Taozhi Wang and Daryl Echeazu and Hubert Yang and Timothy Wu and Spandan Patel and Vidhi Kulkarni and Vijaykaarti Sundarapandiyan and Ashley Zhang and Andrew Le and Zafir Nasim and Srikar Yalam and Ritesh Kasamsetty and Soham Samal and Hubert Yang and David Sun and Nihar Shah and Abhijeet Saha and Alex Zhang and Leon Nguyen and Laasya Nagumalli and Kaixin Wang and Alan Zhou and Aidan Wu and Jason Luo and Anwith Telluri and Summer Yue and Alexandr Wang and Dan Hendrycks},
      year={2025},
      eprint={2501.14249},
      archivePrefix={arXiv},
      primaryClass={cs.LG},
      url={https://arxiv.org/abs/2501.14249}, 
}

@misc{omni-olpc,
      title={Omni-MATH: A Universal Olympiad Level Mathematic Benchmark For Large Language Models}, 
      author={Bofei Gao and Feifan Song and Zhe Yang and Zefan Cai and Yibo Miao and Qingxiu Dong and Lei Li and Chenghao Ma and Liang Chen and Runxin Xu and Zhengyang Tang and Benyou Wang and Daoguang Zan and Shanghaoran Quan and Ge Zhang and Lei Sha and Yichang Zhang and Xuancheng Ren and Tianyu Liu and Baobao Chang},
      year={2024},
      eprint={2410.07985},
      archivePrefix={arXiv},
      primaryClass={cs.CL},
      url={https://arxiv.org/abs/2410.07985}, 
}

@misc{nof1,
  title        = {Alpha Arena — Exploring the Limits of Large Language Models as Quant Traders},
  author       = {{Nof1.ai}},
  year         = {2025},
  howpublished = {\url{https://nof1.ai/blog/TechPost1}},
  note         = {Accessed: 2025-12-10}
}

@misc{metr2025measuring,
  title        = {Measuring AI Ability to Complete Long Tasks},
  author       = {{METR}},
  year         = {2025},
  month        = {03},
  day          = {19},
  howpublished = {METR blog},
  url          = {https://metr.org/blog/2025-03-19-measuring-ai-ability-to-complete-long-tasks/}
}

@inproceedings{mialon2023gaiabenchmarkgeneralai,
  title = {{GAIA}: A Benchmark for General {AI} Assistants},
  author = {Mialon, Grégoire and Fourrier, Clémentine and Swift, Craig and Wolf, Thomas and LeCun, Yann and Scialom, Thomas},
  booktitle = {International Conference on Learning Representations},
  year = {2024},
  eprint = {2311.12983},
  archivePrefix = {arXiv},
  primaryClass = {cs.CL},
  url = {https://proceedings.iclr.cc/paper_files/paper/2024/hash/25ae35b5b1738d80f1f03a8713e405ec-Abstract-Conference.html}
}

@misc{browsecomp,
      title={BrowseComp: A Simple Yet Challenging Benchmark for Browsing Agents}, 
      author={Jason Wei and Zhiqing Sun and Spencer Papay and Scott McKinney and Jeffrey Han and Isa Fulford and Hyung Won Chung and Alex Tachard Passos and William Fedus and Amelia Glaese},
      year={2025},
      eprint={2504.12516},
      archivePrefix={arXiv},
      primaryClass={cs.CL},
      url={https://arxiv.org/abs/2504.12516}, 
}

@inproceedings{webarenarea,
  title = {{WebArena}: A Realistic Web Environment for Building Autonomous Agents},
  author = {Zhou, Shuyan and Xu, Frank F. and Zhu, Hao and Zhou, Xuhui and Lo, Robert and Sridhar, Abishek and Cheng, Xianyi and Ou, Tianyue and Bisk, Yonatan and Fried, Daniel and Alon, Uri and Neubig, Graham},
  booktitle = {International Conference on Learning Representations},
  year = {2024},
  eprint = {2307.13854},
  archivePrefix = {arXiv},
  primaryClass = {cs.AI},
  url = {https://proceedings.iclr.cc/paper_files/paper/2024/hash/4410c0711e9154a7a2d26f9b3816d1ef-Abstract-Conference.html}
}

@inproceedings{mind2web,
  title = {{Mind2Web}: Towards a Generalist Agent for the Web},
  author = {Deng, Xiang and Gu, Yu and Zheng, Boyuan and Chen, Shijie and Stevens, Sam and Wang, Boshi and Sun, Huan and Su, Yu},
  booktitle = {Advances in Neural Information Processing Systems},
  volume = {36},
  year = {2023},
  eprint = {2306.06070},
  archivePrefix = {arXiv},
  primaryClass = {cs.CL},
  url = {https://proceedings.neurips.cc/paper_files/paper/2023/hash/5950bf290a1570ea401bf98882128160-Abstract-Datasets_and_Benchmarks.html}
}

@misc{tbench_2025,
      title={Terminal-Bench: A Benchmark for AI Agents in Terminal Environments}, 
      url={https://github.com/laude-institute/terminal-bench}, 
      author={The Terminal-Bench Team}, year={2025}, month={Apr}}

@article{supplychain,
    author = {Grewal, Dhruv and Nordfält, Jens and Roggeveen, Anne and Olbrich, Rainer and Jansen, Hans Christian},
    title = {Price-quality relationship in pricing strategies for private labels},
    journal = {Journal of Product and Brand Management},
    volume = {23},
    number = {6},
    pages = {429-438},
    year = {2014},
    month = {09},
    issn = {1061-0421},
    doi = {10.1108/JPBM-06-2014-0627},
    url = {https://doi.org/10.1108/JPBM-06-2014-0627},
    eprint = {https://www.emerald.com/jpbm/article-pdf/23/6/429/1581797/jpbm-06-2014-0627.pdf},
}

@inproceedings{reflection,
  title = {Reflexion: Language Agents with Verbal Reinforcement Learning},
  author = {Shinn, Noah and Cassano, Federico and Berman, Edward and Gopinath, Ashwin and Narasimhan, Karthik and Yao, Shunyu},
  booktitle = {Advances in Neural Information Processing Systems},
  volume = {36},
  year = {2023},
  eprint = {2303.11366},
  archivePrefix = {arXiv},
  primaryClass = {cs.AI},
  url = {https://proceedings.neurips.cc/paper_files/paper/2023/hash/1b44b878bb782e6954cd888628510e90-Abstract-Conference.html}
}

@misc{utltabench,
      title={UltraHorizon: Benchmarking Agent Capabilities in Ultra Long-Horizon Scenarios}, 
      author={Haotian Luo and Huaisong Zhang and Xuelin Zhang and Haoyu Wang and Zeyu Qin and Wenjie Lu and Guozheng Ma and Haiying He and Yingsha Xie and Qiyang Zhou and Zixuan Hu and Hongze Mi and Yibo Wang and Naiqiang Tan and Hong Chen and Yi R. Fung and Chun Yuan and Li Shen},
      year={2025},
      eprint={2509.21766},
      archivePrefix={arXiv},
      primaryClass={cs.AI},
      url={https://arxiv.org/abs/2509.21766}, 
}

@misc{herobenchbenchmarklonghorizonplanning,
      title={HeroBench: A Benchmark for Long-Horizon Planning and Structured Reasoning in Virtual Worlds}, 
      author={Petr Anokhin and Roman Khalikov and Stefan Rebrikov and Viktor Volkov and Artyom Sorokin and Vincent Bissonnette},
      year={2025},
      eprint={2508.12782},
      archivePrefix={arXiv},
      primaryClass={cs.AI},
      url={https://arxiv.org/abs/2508.12782}, 
}

@misc{odysseybenchevaluatingllmagents,
      title={OdysseyBench: Evaluating LLM Agents on Long-Horizon Complex Office Application Workflows}, 
      author={Weixuan Wang and Dongge Han and Daniel Madrigal Diaz and Jin Xu and Victor Rühle and Saravan Rajmohan},
      year={2025},
      eprint={2508.09124},
      archivePrefix={arXiv},
      primaryClass={cs.CL},
      url={https://arxiv.org/abs/2508.09124}, 
}

@misc{zheng2025lifelongagentbench,
      title={LifelongAgentBench: Evaluating LLM Agents as Lifelong Learners},
      author={Junhao Zheng and Xidi Cai and Qiuke Li and Duzhen Zhang and ZhongZhi Li and Yingying Zhang and Le Song and Qianli Ma},
      year={2025},
      eprint={2505.11942},
      archivePrefix={arXiv},
      primaryClass={cs.AI},
      url={https://arxiv.org/abs/2505.11942},
}

@misc{xu2024theagentcompany,
      title={TheAgentCompany: Benchmarking LLM Agents on Consequential Real World Tasks},
      author={Frank F. Xu and Yufan Song and Boxuan Li and Yuxuan Tang and Kritanjali Jain and Mengxue Bao and Zora Z. Wang and Xuhui Zhou and Zhitong Guo and Murong Cao and Mingyang Yang and Hao Yang Lu and Amaad Martin and Zhe Su and Leander Maben and Raj Mehta and Wayne Chi and Lawrence Jang and Yiqing Xie and Shuyan Zhou and Graham Neubig},
      year={2024},
      eprint={2412.14161},
      archivePrefix={arXiv},
      primaryClass={cs.AI},
      url={https://arxiv.org/abs/2412.14161},
}

@inproceedings{3webshopscalablerealworldweb,
  title = {{WebShop}: Towards Scalable Real-World Web Interaction with Grounded Language Agents},
  author = {Yao, Shunyu and Chen, Howard and Yang, John and Narasimhan, Karthik},
  booktitle = {Advances in Neural Information Processing Systems},
  volume = {35},
  year = {2022},
  eprint = {2207.01206},
  archivePrefix = {arXiv},
  primaryClass = {cs.CL},
  url = {https://proceedings.neurips.cc/paper_files/paper/2022/hash/82ad13ec01f9fe44c01cb91814fd7b8c-Abstract-Conference.html}
}

@misc{erdogan2025planandactimprovingplanningagents,
      title={Plan-and-Act: Improving Planning of Agents for Long-Horizon Tasks}, 
      author={Lutfi Eren Erdogan and Nicholas Lee and Sehoon Kim and Suhong Moon and Hiroki Furuta and Gopala Anumanchipalli and Kurt Keutzer and Amir Gholami},
      year={2025},
      eprint={2503.09572},
      archivePrefix={arXiv},
      primaryClass={cs.CL},
      url={https://arxiv.org/abs/2503.09572}, 
}

@inproceedings{OFCOURSE,
 author = {Zhu, Yiheng and Zhan, Yang and Huang, Xuankun and Chen, Yuwei and Chen, yujie and Wei, Jiangwen and Feng, Wei and Zhou, Yinzhi and Hu, Haoyuan and Ye, Jieping},
 booktitle = {Advances in Neural Information Processing Systems},
 editor = {A. Oh and T. Naumann and A. Globerson and K. Saenko and M. Hardt and S. Levine},
 pages = {34765--34777},
 publisher = {Curran Associates, Inc.},
 title = {OFCOURSE: A Multi-Agent Reinforcement Learning Environment for Order Fulfillment},
 url = {https://proceedings.neurips.cc/paper_files/paper/2023/file/6d0cfc5db3feeabf6762129ba91bd3a1-Paper-Datasets_and_Benchmarks.pdf},
 volume = {36},
 year = {2023}
}

@misc{leluc2023marlimmultiagentreinforcementlearning,
      title={MARLIM: Multi-Agent Reinforcement Learning for Inventory Management}, 
      author={Rémi Leluc and Elie Kadoche and Antoine Bertoncello and Sébastien Gourvénec},
      year={2023},
      eprint={2308.01649},
      archivePrefix={arXiv},
      primaryClass={cs.LG},
      url={https://arxiv.org/abs/2308.01649}, 
}

@misc{aimbenchevaluatingdecisionmakingbiases,
      title={AIM-Bench: Evaluating Decision-making Biases of Agentic LLM as Inventory Manager}, 
      author={Xuhua Zhao and Yuxuan Xie and Caihua Chen and Yuxiang Sun},
      year={2025},
      eprint={2508.11416},
      archivePrefix={arXiv},
      primaryClass={cs.AI},
      url={https://arxiv.org/abs/2508.11416}, 
}

@misc{invagentlargelanguagemodel,
      title={InvAgent: A Large Language Model based Multi-Agent System for Inventory Management in Supply Chains}, 
      author={Yinzhu Quan and Zefang Liu},
      year={2025},
      eprint={2407.11384},
      archivePrefix={arXiv},
      primaryClass={cs.CL},
      url={https://arxiv.org/abs/2407.11384}, 
}

@inproceedings{inventory,
  author = {Shar, Ibrahim and Sun, Wenhuan and Wang, Haiyan and Gupta, Chetan},
  title = {Deep Reinforcement Learning toward Robust Multi-Echelon Supply Chain Inventory Optimization},
  booktitle = {2022 IEEE 18th International Conference on Automation Science and Engineering (CASE)},
  year = {2022},
  month = aug,
  pages = {1385--1391},
  doi = {10.1109/CASE49997.2022.9926659}
}

@misc{xia2023retailsynth,
      title={RetailSynth: Synthetic Data Generation for Retail AI Systems Evaluation},
      author={Yu Xia and Ali Arian and Sriram Narayanamoorthy and Joshua Mabry},
      year={2023},
      eprint={2312.14095},
      archivePrefix={arXiv},
      primaryClass={cs.LG},
      url={https://arxiv.org/abs/2312.14095}
}

@misc{xia2024simulationretailpromotions,
      title={Simulation-Based Benchmarking of Reinforcement Learning Agents for Personalized Retail Promotions},
      author={Yu Xia and Sriram Narayanamoorthy and Zhengyuan Zhou and Joshua Mabry},
      year={2024},
      eprint={2405.10469},
      archivePrefix={arXiv},
      primaryClass={cs.LG},
      url={https://arxiv.org/abs/2405.10469}
}

@misc{wang2025freshretailnet,
      title={FreshRetailNet-50K: A Stockout-Annotated Censored Demand Dataset for Latent Demand Recovery and Forecasting in Fresh Retail},
      author={Yangyang Wang and Jiawei Gu and Li Long and Xin Li and Li Shen and Zhouyu Fu and Xiangjun Zhou and Xu Jiang},
      year={2025},
      eprint={2505.16319},
      archivePrefix={arXiv},
      primaryClass={cs.LG},
      url={https://arxiv.org/abs/2505.16319}
}

@misc{tkachuk2024consumertransactions,
      title={Consumer Transactions Simulation through Generative Adversarial Networks},
      author={Sergiy Tkachuk and Szymon Lukasik and Anna Wroblewska},
      year={2024},
      eprint={2408.03655},
      archivePrefix={arXiv},
      primaryClass={cs.LG},
      url={https://arxiv.org/abs/2408.03655}
}

@inproceedings{wang2025shoppingbench,
  title = {{ShoppingBench}: A Real-World Intent-Grounded Shopping Benchmark for {LLM}-based Agents},
  author = {Wang, Jiangyuan and Xiao, Kejun and Sun, Qi and Zhao, Huaipeng and Luo, Tao and Zhang, Jian Dong and Zeng, Xiaoyi},
  booktitle = {Proceedings of the AAAI Conference on Artificial Intelligence},
  volume = {40},
  pages = {33521--33529},
  year = {2026},
  doi = {10.1609/aaai.v40i39.40640},
  eprint = {2508.04266},
  archivePrefix = {arXiv},
  primaryClass = {cs.CL},
  url = {https://ojs.aaai.org/index.php/AAAI/article/view/40640}
}

@inproceedings{yao2023react,
  title = {{ReAct}: Synergizing Reasoning and Acting in Language Models},
  author = {Yao, Shunyu and Zhao, Jeffrey and Yu, Dian and Du, Nan and Shafran, Izhak and Narasimhan, Karthik and Cao, Yuan},
  booktitle = {International Conference on Learning Representations},
  year = {2023},
  eprint = {2210.03629},
  archivePrefix = {arXiv},
  primaryClass = {cs.CL},
  url = {https://openreview.net/forum?id=WE_vluYUL-X}
}

\appendix
\section{Environment Configuration Details}
\label{appendix:env_details}

\begin{table*}[t]
\centering
\footnotesize
\setlength{\tabcolsep}{2.3pt}
\renewcommand{\arraystretch}{1.08}
\newcommand{\benchref}[2]{\parbox[t]{3.35cm}{#1\\[-1pt]{\scriptsize #2}}}
\begin{tabular}{@{}p{3.35cm}c c c*{7}{c}@{}}
\toprule
Benchmark & \shortstack{Decision\\maker} & \shortstack{Data\\grounded} & Horizon & \shortstack{Price--\\Inv.} & \shortstack{Perish./\\Stockout} & Shelf & Reviews & Supplier & News & Finance \\
\midrule
\benchref{RetailSynth}{\citep{xia2023retailsynth}} & Policy & \xmark & Medium & \cmark & \xmark & \xmark & \xmark & \xmark & \xmark & \xmark \\
\benchref{Promo-RL}{\citep{xia2024simulationretailpromotions}} & Policy & \xmark & Medium & \cmark & \xmark & \xmark & \xmark & \xmark & \xmark & \xmark \\
\benchref{FreshRetailNet}{\citep{wang2025freshretailnet}} & None & \cmark & Static & \xmark & \cmark & \xmark & \xmark & \xmark & \xmark & \xmark \\
\benchref{Transaction sim.}{\citep{tkachuk2024consumertransactions}} & None & \cmark & Static & \xmark & \cmark & \xmark & \xmark & \xmark & \xmark & \xmark \\
\benchref{ShoppingBench}{\citep{wang2025shoppingbench}} & Agents & \cmark & Medium & \xmark & \xmark & \xmark & \cmark & \xmark & \xmark & \xmark \\
\benchref{WebShop}{\citep{3webshopscalablerealworldweb}} & Agents & \cmark & Medium & \xmark & \xmark & \xmark & \cmark & \xmark & \xmark & \xmark \\
\benchref{OFCOURSE}{\citep{OFCOURSE}} & Policy & \xmark & Long & \xmark & \cmark & \xmark & \xmark & \xmark & \xmark & \xmark \\
\benchref{Inventory opt.}{\citep{inventory}} & Policy & \xmark & Long & \xmark & \cmark & \xmark & \xmark & \xmark & \xmark & \xmark \\
\benchref{MARLIM}{\citep{leluc2023marlimmultiagentreinforcementlearning}} & Policy & \xmark & Long & \xmark & \cmark & \xmark & \xmark & \xmark & \xmark & \xmark \\
\benchref{InvAgent}{\citep{invagentlargelanguagemodel}} & Agents & \xmark & Medium & \xmark & \cmark & \xmark & \xmark & \xmark & \xmark & \xmark \\
\benchref{AIM-Bench}{\citep{aimbenchevaluatingdecisionmakingbiases}} & Agents & \xmark & Medium & \xmark & \cmark & \xmark & \xmark & \xmark & \xmark & \xmark \\
\benchref{VendingBench}{\citep{backlund2025vendingbenchbenchmarklongtermcoherence,andonlabs2025vendingbench2}} & Agents & \xmark & Long & \cmark & \xmark & \xmark & \xmark & \cmark & \xmark & \cmark \\
\textbf{RetailBench} & \textbf{Agents} & \cmark & \textbf{Long} & \cmark & \cmark & \cmark & \cmark & \cmark & \cmark & \cmark \\
\bottomrule
\end{tabular}
\caption{Comparison with related retail, supply-chain, e-commerce, and long-horizon agent benchmarks. The table emphasizes the decision surface exposed to the decision maker. ``Agents'' denotes LLM/tool-using agents, ``Policy'' denotes learned optimization or RL policies, and ``None'' denotes datasets or simulators without an autonomous decision maker. A checkmark indicates that the corresponding axis is explicitly modeled as an environment state, action, or evaluation target.}
\label{tab:benchmark_axes}
\end{table*}

\subsection{Simulator Modules}
\label{app:simulator_modules}

RetailBench follows the seven-module organization shown in Figure~\ref{fig:retailbench_environment}: Inventory, Suppliers \& Orders, Shelf, Customer Demand, Reviews \& Returns, News Events, and Finance \& Records. Static product attributes, including identity, category, shelf life, retail price, historical sales summaries, and review summaries, are shared across these modules. The modules are coupled through the daily transition process: orders create delayed deliveries, shelf-assortment decisions determine customer-facing availability, demand is realized as sales, reviews and returns update feedback signals, news changes demand or supply conditions, and finance records cash-flow outcomes and store viability.

\paragraph{Inventory.}
The inventory module tracks physical stock state, including on-hand units, item age, sellable quantity, capacity constraints, stockouts, expirations, and waiting items that cannot yet enter inventory because of capacity limits. It determines the feasibility constraints for procurement, shelving, and sales.

\paragraph{Shelf.}
The shelf module enforces shelf-capacity constraints and determines the customer-facing assortment. At each day $t$, it maintains an active assortment $V_t \subseteq \mathcal{J}$ subject to the capacity constraint $|V_t| \le C^{\mathrm{shelf}}$. Stocked products outside $V_t$ remain in inventory but are not visible to customers and therefore cannot generate demand.

\paragraph{Suppliers \& Orders.}
The suppliers-and-orders module provides five procurement candidates per product, each with cost, quality, and delivery-time profiles. Costs are grounded in Dominick's, and quality tiers follow documented price--quality relationships in private-label retailing~\citep{supplychain}; supplier choices affect procurement expenditure, delivery timing, return risk, reviews, and future inventory availability. Replenishment orders are converted into delayed deliveries that enter inventory on later days when capacity allows.

\paragraph{Customer Demand.}
Following RetailSynth~\citep{xia2023retailsynth}, customer choice follows an MNL model~\citep{mcfadden_conditional_1974}. Daily traffic comes from Dominick's~\citep{DominicksDataset}. The simulator first samples category demand from power-aggregated shelf-visible attraction~\citep{nakanishi1974mci}, then allocates it across visible products. The corresponding update equations are given in Appendix~\ref{app:simulator_update_equations}.

\paragraph{Reviews \& Returns.}
The reviews-and-returns module converts realized sales and supplier quality into post-sale feedback. Sold items can generate product-level reviews, which update rating summaries and later demand through $\rho^{\mathrm{rev}}_{jt}$. Returned units are sampled from supplier-quality-dependent return probabilities, logged at the product and supplier level, and deducted from realized revenue. Thus, supplier quality affects both customer feedback and return pressure.

\paragraph{News Events.}
The news-events module samples observable news text with hidden simulator metadata specifying impact scope, target category or product, direction, strength, affected side, and duration. Demand-side news changes product attraction through $\rho^{\mathrm{news}}_{jt}$, while supply-side news changes supplier costs through $\delta^{\mathrm{sup}}_{jt}$.

\paragraph{Finance \& Records.}
The finance-and-records module tracks revenue, procurement cost, returns, rent, cash, inventory value, pending-order value, net worth, and persistent sales, supplier, review, return, and news records. At each day transition, the simulator applies deliveries, shelf-constrained demand realization, sales, feedback updates, expiration, rent, cash-flow updates, and next-day supplier/news changes.

\subsection{State and Action Spaces}
\label{app:state_action_spaces}

We construct the environment using real-world retail data from the Dominick’s dataset \citep{DominicksDataset}. From the 20 available product categories, we select 96 products with the most complete and informative sales records. The richness of these observations enables reliable modeling of the relationship between pricing decisions and realized demand.

\paragraph{Key Symbols.}
Let $\mathcal{J}$ denote the set of selected products, with cardinality $|\mathcal{J}| = J$ (where $J=96$ in our experiments). Each product $j \in \mathcal{J}$ belongs to a product category $\mathrm{cat}(j) \in \{1,\ldots,20\}$. For each product $j$, let $\mathcal{K}(j)$ denote its associated supplier set, with $|\mathcal{K}(j)| = 5$. Let $V_t\subseteq\mathcal{J}$ denote the product set placed on shelf on day $t$.

\paragraph{State and action spaces.}
The simulator state space is $\mathcal{S}=\mathcal{S}^{\mathrm{prod}}\times\mathcal{S}^{\mathrm{inv}}\times\mathcal{S}^{\mathrm{shelf}}\times\mathcal{S}^{\mathrm{sup}}\times\mathcal{S}^{\mathrm{dem}}\times\mathcal{S}^{\mathrm{ext}}\times\mathcal{S}^{\mathrm{fin}}$, covering product attributes and prices, inventory and pending deliveries, shelf assortment, supplier candidates, demand signals, external news, and financial accounting. The action space is $\mathcal{A}=\mathcal{A}^{\mathrm{read}}\cup\mathcal{A}^{\mathrm{other}}\cup\mathcal{A}^{\mathrm{act}}$: $\mathcal{A}^{\mathrm{read}}$ contains observation tools for funds, inventory, shelf status, product prices, costs, sales/profit history, supplier quotes, notes, review-rating summaries, supplier-return-rate summaries, and news; $\mathcal{A}^{\mathrm{other}}$ contains constrained analysis tools such as \texttt{execute\_code}; and $\mathcal{A}^{\mathrm{act}}$ contains \texttt{place\_order}, \texttt{modify\_product\_price}, \texttt{set\_shelf\_products}, \texttt{add\_note}, \texttt{remove\_note}, and \texttt{end\_today}.

\subsubsection{Product State \texorpdfstring{$S_t^{\text{prod}}$}{S t prod}}

For each product $j$, we maintain a set of static attributes together with time-varying operational signals:
\begin{equation}
S_{t,j}^{\text{prod}} =
\big(
\texttt{id}_j,\ \texttt{desc}_j,\ L_j,\ p_{jt},\ \mathbf{h}^{\text{sales}}_{j,t},\ \mathbf{r}_{j,t}
\big).
\end{equation}
Here, $\texttt{id}_j$ and $\texttt{desc}_j$ denote the unique identifier and textual description of product $j$, respectively. $L_j$ denotes the shelf life (in days), and $p_{jt}$ is the retail price on day $t$. The term $\mathbf{h}^{\text{sales}}_{j,t}$ encodes historical sales statistics, while $\mathbf{r}_{j,t}$ summarizes aggregated customer review signals, such as average rating and review volume.

\paragraph{Data grounding.}
The product set and cost priors are constructed from the Dominick’s dataset \citep{DominicksDataset}. Historical sales records from the same source are used to calibrate the demand model.

\subsubsection{Inventory State \texorpdfstring{$S_t^{\text{inv}}$}{S t inv}}

Inventory is represented at the product level with age tracking. Let $I_{jt}$ denote the on-hand units of product $j$ at the beginning of day $t$. To support shelf-life constraints and depreciation, we track the arrival time and remaining shelf life of each unit.

\paragraph{Capacity constraint and pending queue.}
Let $\mathrm{Cap}$ denote the total inventory capacity of the store. If incoming replenishment orders would violate this constraint, excess units are placed into a first-in-first-out (FIFO) pending queue $\mathcal{Q}_t$ and become available only after sufficient inventory space is released:
\begin{equation}
\sum_{j\in\mathcal{J}} \sum_{a=0}^{L_j} I_{jt}^{(a)} \le \mathrm{Cap}.
\end{equation}

\paragraph{Stockout signal.}
When realized demand exceeds the available sellable inventory of product $j$ on day $t$, a stockout indicator is triggered. This signal is recorded at the end of day $t$ and exposed to the agent as explicit operational feedback.

\paragraph{Expiration and destruction.}
At each day transition, inventory units whose residence time exceeds their shelf life are removed from the system.

\paragraph{Shelf assortment.}
RetailBench additionally tracks an active shelf assortment $V_t$. The shelf has finite product capacity:
\begin{equation}
V_t \subseteq \mathcal{J}, \qquad |V_t| \le C^{\mathrm{shelf}}.
\end{equation}
Only products in $V_t$ are visible to customer demand; newly delivered inventory remains off shelf until the agent updates $V_t$ through the shelf-assortment action.

\subsubsection{Supply Chain State \texorpdfstring{$S_t^{\text{sup}}$}{S t sup}}

Each product $j$ is associated with a set of suppliers $k \in \mathcal{K}(j)$. The supplier state includes procurement price $c_{jk}$, quality level $q_{jk}$, and delivery-time distribution:
\begin{equation}
S_{t,j}^{\text{sup}} =
\big\{(c_{jk}, q_{jk}, \mathcal{L}_{jk}) : k \in \mathcal{K}(j)\big\}.
\end{equation}
Delivery time $\ell_{jk} \sim \mathcal{L}_{jk}$ is sampled when an order is placed. Procurement prices are discretized into tiers using Dominick’s cost signals \citep{DominicksDataset}, following empirically observed positive correlations between price tier and quality level \citep{supplychain}.

\paragraph{Enforced diversity.}
To avoid degenerate supplier configurations, we enforce that each product has (i) one supplier with maximal quality, (ii) one supplier with minimal price and minimal quality, and remaining suppliers spanning intermediate price–quality levels.

\subsubsection{Demand Signals and Demand Generation}
\label{app:demand}

Demand-related components capture customer traffic, shelf-visible choice sets, and stochastic sales generation. We define
\begin{equation}
S_t^{\text{dem}} =
\big(
N_t,\ \{A^0_{jt}\}_{j\in\mathcal{J}},\
\{\rho^{\mathrm{rev}}_{jt}\}_{j\in\mathcal{J}},\
\{\rho^{\mathrm{news}}_{jt}\}_{j\in\mathcal{J}}
\big),
\end{equation}
where $N_t$ denotes daily customer traffic, $A^0_{jt}$ is the calibrated base attraction, and $\rho^{\mathrm{rev}}_{jt}$ and $\rho^{\mathrm{news}}_{jt}$ are bounded demand-side effects from reviews and news.

\paragraph{Customer traffic.}
Daily customer traffic is derived from the Dominick’s dataset~\citep{DominicksDataset}.

\paragraph{Consumer choice and demand generation.}
Given traffic $N_t$, prices $\{p_{jt}\}_{j\in\mathcal{J}}$, shelf assortment $V_t$, review signals, and active news, RetailBench uses the attraction model in Equations~\ref{eq:base_attraction}--\ref{eq:demand_sales}. The current implementation aggregates shelf-visible product attractions within each category using a power mean, samples category-level demand against an outside option, allocates category demand with a multinomial draw over relative product attractions, and caps realized sales by sellable inventory. If $j\notin V_t$, then product $j$ cannot receive demand on that day.

\subsubsection{External Information \texorpdfstring{$S_t^{\text{ext}}$}{S t ext} (News Module)}

We maintain a set of active news events $\mathcal{E}_t$. Each event $e \in \mathcal{E}_t$ is represented as
\begin{equation}
\begin{aligned}
e = (&\texttt{type},\ \texttt{scope},\ \texttt{target},\\
     &\texttt{side},\ \texttt{sign},\ \eta,\\
     &\texttt{text},\ \texttt{ttl})
\end{aligned}
\end{equation}

where \texttt{scope} $\in \{\text{macro}, \text{category}, \text{product}, \text{neutral}\}$,
\texttt{side} $\in \{\text{demand}, \text{supply}, \text{both}\}$,
\texttt{sign} $\in \{+1,-1\}$,
$\eta$ denotes impact magnitude, and \texttt{ttl} is the time-to-live in days. News texts are grounded in financial-news-articles~\citep{ashraq2025financialNewsArticles} and expanded into retail-relevant events through LLM-assisted rewriting and metadata annotation.

\paragraph{Impact application.}
News impacts are incorporated into (i) demand utilities and/or (ii) supplier prices depending on \texttt{side} and \texttt{scope}. Neutral news has $\eta = 0$ by design.

\subsubsection{Simulator Update Equations}
\label{app:simulator_update_equations}

This section gives the detailed equations summarized in Section~\ref{sec:module_logic}.

\paragraph{Supply price update.}
Let $\bar c_{jkt}$ be the base procurement price for supplier $k$ and product $j$ on day $t$, and let $\delta^{\mathrm{sup}}_{jt}$ be the active supply-side news effect matched to product $j$, its category, or the macro scope. The daily procurement price exposed to the agent is
\begin{equation}
c_{jkt}=\bar c_{jkt}\bigl(1-\delta^{\mathrm{sup}}_{jt}\bigr).
\label{eq:supply_price}
\end{equation}
Positive supply news lowers procurement cost, while negative supply news raises it.

\paragraph{Attraction and sales.}
The base product attraction combines historical preference, seasonality, price sensitivity, and stochastic demand noise:
\begin{equation}
\begin{aligned}
A^{0}_{jt}
=\exp\!\Big(&\alpha_j+
\bigl(\beta_j+\beta^s_j\sin(2\pi w_t/52)\\
&+\beta^c_j\cos(2\pi w_t/52)\bigr)p_{jt}
+\epsilon_{jt}\Big).
\end{aligned}
\label{eq:base_attraction}
\end{equation}
Reviews and news multiplicatively adjust this attraction:
\begin{equation}
A_{jt}=A^{0}_{jt}\bigl(1+\rho^{\mathrm{rev}}_{jt}+\rho^{\mathrm{news}}_{jt}\bigr).
\label{eq:adjusted_attraction}
\end{equation}
Let $V_{ct}=\{j\in V_t:\mathrm{cat}(j)=c\}$ be the shelf-visible product set in category $c$. Given daily traffic $N_t$, RetailBench samples category-level purchase volume and allocates it by relative attraction among shelf-visible products:
\begin{equation}
\begin{aligned}
G_{ct}
&=\left(\sum_{j\in V_{ct}} A_{jt}^{\rho}\right)^{1/\rho},\\
\tilde Y_{ct}
&\sim \mathrm{Binomial}\!\left(N_t,\frac{G_{ct}}{1+G_{ct}}\right),\\
\pi_{jt|c}
&=\frac{A_{jt}}{\sum_{i\in V_{ct}}A_{it}}, \quad j\in V_{ct},\\
\hat{\mathbf{y}}_{ct}
&\sim \mathrm{Multinomial}\!\left(\tilde Y_{ct},\boldsymbol{\pi}_{ct}\right),\\
y_{jt}
&=\min\!\left(\hat y_{jt}, I_{jt}^{\mathrm{sellable}}\right).
\end{aligned}
\label{eq:demand_sales}
\end{equation}

\paragraph{Review effect.}
Review text is grounded in Amazon Reviews 2023~\citep{hou2024bridging}. During simulation, sold units can generate reviews; the rating is sampled from supplier quality and the text is sampled from the matching category--rating--dimension pool. The demand effect is computed from the difference between recent and initial rating:
\begin{equation}
\rho^{\mathrm{rev}}_{jt}
=g_{\mathrm{cat}(j)}(\bar r_{jt})-g_{\mathrm{cat}(j)}(r^{0}_{j}),
\label{eq:review_effect}
\end{equation}
where $g_{\mathrm{cat}(j)}(\cdot)$ is a category-specific smooth rating-gain function, $\bar r_{jt}$ is the recent average rating, and $r^{0}_{j}$ is the initial rating.

\paragraph{News effect.}
Each news event is stored as
\begin{equation}
e=(m,\tau,\texttt{side},\texttt{direction},s,\texttt{text},\texttt{ttl}),
\label{eq:news_record}
\end{equation}
where $m$ is event mode, $\tau$ is target scope, \texttt{side} specifies demand, supply, or both, $s$ is impact strength, and \texttt{ttl} controls how long the event remains active. Demand-side news contributes
\begin{equation}
\rho^{\mathrm{news}}_{jt}
=\lambda\sum_{e\in\mathcal{E}_t:\,e\sim j}
w_{m(e)}\,\mathrm{sgn}(e)\,s_e,
\label{eq:news_effect}
\end{equation}
where $e\sim j$ indicates that the event matches the product, its category, or the macro scope; $w_{m(e)}$ is the mode weight; and $\lambda$ is the global impact scale. Supply-side events analogously define $\delta^{\mathrm{sup}}_{jt}$ in Equation~\ref{eq:supply_price}.

\subsubsection{Financial State \texorpdfstring{$S_t^{\text{fin}}$}{S t fin}}

Let $F_t$ denote available funds at the start of day $t$. Net worth is computed as
\begin{equation}
\mathrm{NW}_t = F_t + \sum_{j\in\mathcal{J}} \sum_{a=0}^{L_j} I_{jt}^{(a)} \cdot v_j(a),
\end{equation}
where $v_j(a)$ denotes the per-unit value at age $a$. We adopt linear shelf-life depreciation:
\begin{equation}
v_j(a) = c_j \cdot \max\!\left(0,\ 1 - \frac{a}{L_j}\right),
\end{equation}
with $c_j$ denoting a reference procurement cost (e.g., the mean or realized supplier cost).

\subsubsection{Action Space Details}
\label{app:action_space_details}

RetailBench exposes actions through typed tools. Read-only tools instantiate $\Omega$ and do not modify simulator state; other tools provide constrained analysis over observations; state-changing tools instantiate $T$.

\paragraph{Read-only tools.}
\begin{itemize}[leftmargin=*,nosep]
    \item \texttt{view\_funds\_and\_date}: returns current funds, date, and rent context.
    \item \texttt{view\_inventory}: reports on-hand units, waiting-capacity items, incoming quantities, and open supplier orders.
    \item \texttt{view\_shelf\_status}: reports active shelf assortment, shelf capacity, and on-shelf product summaries.
    \item \texttt{view\_product\_prices}: returns current selling prices for selected products.
    \item \texttt{view\_product\_inventory\_cost}: summarizes average procurement cost and inventory age.
    \item \texttt{view\_sales\_profit\_history}: returns historical sales, average selling price, and realized profit.
    \item \texttt{view\_current\_date\_supplier\_prices}: returns current supplier quotes.
    \item \texttt{view\_supplier\_price\_history}: returns historical supplier prices for a product.
    \item \texttt{view\_notes}: returns the agent's stored cross-day notes.
\end{itemize}

\paragraph{Conditional observation tools.}
\begin{itemize}[leftmargin=*,nosep]
    \item \texttt{view\_product\_avg\_ratings}: exposes product-level average ratings when review feedback is enabled.
    \item \texttt{view\_supplier\_returns\_avg\_rate}: exposes supplier-level return rates, optionally restricted to selected products.
    \item \texttt{view\_today\_news}: lists current news IDs and titles when news is enabled.
    \item \texttt{view\_news\_detail}: returns the observable content of a selected news item.
    \item \texttt{view\_news\_history}: returns news records over a date range.
\end{itemize}

\paragraph{Other/analysis tools.}
\begin{itemize}[leftmargin=*,nosep]
    \item \texttt{execute\_code}: runs constrained Python for read-only analysis over observations. It can compute summaries, rank products, and compare suppliers, but mutation tools are unavailable inside the sandbox.
\end{itemize}

\paragraph{State-changing tools.}
\begin{itemize}[leftmargin=*,nosep]
    \item \texttt{place\_order}: places a multi-product order through a selected supplier, deducts procurement cost, samples delivery time, and schedules future inventory arrivals.
    \item \texttt{modify\_product\_price}: updates the retail price $p_{jt}$ of product $j$.
    \item \texttt{set\_shelf\_products}: replaces the active shelf assortment subject to $|V'|\le C^{\mathrm{shelf}}$.
    \item \texttt{add\_note} and \texttt{remove\_note}: update cross-day notes.
    \item \texttt{end\_today}: applies $T_{\mathrm{day}}$ over deliveries, shelf-visible demand, sales, reviews, returns, expiration, rent, cash-flow updates, and next-day supplier/news changes.
\end{itemize}

\subsection{Environment Setting}
\label{appendix:env_setting}

RetailBench provides a single released environment configuration for all reported evaluations.

\begin{itemize}
    \item \textbf{Time Range:}
    \begin{itemize}
        \item Data begin time: 06/06/91
        \item Data end time: 12/31/95
        \item Store begin time: 09/07/91
        \item Store ID: 15
    \end{itemize}
    
    \item \textbf{Financial Parameters:}
    \begin{itemize}
        \item Initial funds: 30,000
        \item Daily rent: 600
        \item Inventory capacity: 15,000
        \item Shelf capacity: 40 products
    \end{itemize}
    
    \item \textbf{Feature Enablement:}
    \begin{itemize}
        \item Review enabled: \texttt{True}
        \item Review ratio: 0.05
        \item News enabled: \texttt{True}
        \item Shelf constraint enabled: \texttt{True}
    \end{itemize}
    
    \item \textbf{News Configuration:}
    \begin{itemize}
        \item News impact base scale: 0.4
        \item News daily count: 20
        \item News random seed: 42
        \item News sample ratios:
        \begin{itemize}
            \item Neutral: 0.9
            \item Single category: 0.02
            \item Macro all: 0.03
            \item product level: 0.05
        \end{itemize}
        \item News impact mode weights:
        \begin{itemize}
            \item Neutral: 0.0
            \item Macro all: 1.0
            \item Single category: 1.0
            \item product level: 1.2
        \end{itemize}
    \end{itemize}
    
    \item \textbf{Selected Categories (20 categories):}
    \begin{itemize}
        \item Bathroom\_Tissues, Beer, Bottled\_Juices, Canned\_Soup, Canned\_Tuna
        \item Cereals, Cheeses, Cigarettes, Cookies, Crackers
        \item Dish\_Detergent, Fabric\_Softeners, Front\_end\_candies, Frozen\_Entrees
        \item Frozen\_Juices, Oatmeal, Paper\_Towels, Snack\_Crackers
        \item Soft\_Drinks, Toothpastes
    \end{itemize}
    
    \item \textbf{Category Effects:} All categories have a uniform effect of -0.1
    
    \item \textbf{Random Seed:} 42
\end{itemize}

\subsection{Oracle Policy}
\label{appendix:handcrafted_policy}

The oracle policy is a deterministic, quality-based store-operation rule used as a privileged non-agent reference. It is not an optimal policy and should not be interpreted as a language-agent baseline: it has direct access to structured simulator fields such as supplier quality scores and news impact values, and it does not face prompt length, tool-selection, memory, or action-formatting constraints. The reported run uses the released environment with $n_{\mathrm{cat}}=2$ products per category, shelf capacity $C^{\mathrm{shelf}}=40$, and replenishment multiplier $B=6$, selected by a parameter sweep over the shelf-constrained setting.

\begin{algorithm}[t]
\small
\caption{Oracle quality-based policy}
\label{alg:handcrafted_policy}
\begin{algorithmic}[1]
\Require Horizon $T$, categories $\mathcal{C}$, products $\mathcal{J}$, shelf capacity $C^{\mathrm{shelf}}$, per-category product count $n_{\mathrm{cat}}$, bulk multiplier $B$.
\State Select $\mathcal{U}\leftarrow\emptyset$.
\For{each category $c\in\mathcal{C}$}
    \State Rank products in $c$ by historical total profit before the store start date.
    \State Add the top $n_{\mathrm{cat}}$ products to $\mathcal{U}$.
\EndFor
\State Set shelf assortment $V_0$ to a category-diverse subset of $\mathcal{U}$ with $|V_0|\le C^{\mathrm{shelf}}$.
\For{day $t=1,\ldots,T$}
    \If{funds are negative for 10 consecutive days}
        \State Terminate the rollout.
    \EndIf
    \State Observe funds, news, inventory, shelf status, recent sales, and supplier quotes.
    \For{each product $j\in\mathcal{U}$}
        \State $I_j \leftarrow$ on-hand units plus waiting-capacity units for product $j$.
        \State Choose supplier $k^\star=\arg\max_k(q_{jk},-c_{jkt})$.
        \State Estimate price $p^\star_j$ by maximizing mean 30-day historical profit:
        \Statex \hspace{\algorithmicindent} $p^\star_j=\arg\max_p \operatorname{mean}\{(p-c_{jk^\star t})\,y_{j\tau}:p_{j\tau}=p\}$.
        \State Let $\hat d_j$ be the mean historical sales at $p^\star_j$; use $\hat d_j=10$ if no record is available.
        \State Compute news multiplier $m_j=\operatorname{clip}((1+\eta^{\mathrm{need}}_j)(1+\eta^{\mathrm{sup}}_j),0.5,2.0)$.
        \State Set target demand $\tilde d_j \leftarrow m_j\hat d_j$.
        \If{$p^\star_j$ exists}
            \State Call \texttt{modify\_product\_price}$(j,p^\star_j)$.
        \EndIf
        \State Let $Q_j$ be the pending order quantity for product $j$.
        \State Let $R_j\leftarrow B\tilde d_j$ and $O_j\leftarrow \max(1,\lfloor B\tilde d_j\rfloor)$.
        \If{$I_j\le R_j$ and $Q_j<O_j$}
            \State $a_j\leftarrow \min(O_j-Q_j,\ \text{available inventory capacity}-100)$.
            \If{$a_j>0$}
                \State Call \texttt{place\_order} for $a_j$ units of product $j$ from supplier $k^\star$.
            \EndIf
        \EndIf
    \EndFor
    \State Call \texttt{end\_today}.
\EndFor
\end{algorithmic}
\end{algorithm}

\subsection{Four-Stage Diagnostic Metrics}
\label{appendix:stage_metrics}

This section defines the stage-level diagnostics used in Section~\ref{sec:failure_pattern_analysis}. The metrics are computed after each rollout by aligning three sources: tool-call traces, simulator records, and realized delayed events. They are diagnostic rather than reward functions. In particular, some quantities use retrospective information or hidden simulator fields to explain agent behavior; they should not be interpreted as information available to the agent at decision time.

We evaluate the daily operational loop in four stages. Stage 1 asks whether important products entered the agent's action set. Stage 2 asks whether the agent gathered the evidence needed before acting. Stage 3 asks whether the gathered evidence was converted into quality-adjusted prices and supplier choices. Stage 4 asks whether the agent later revisited products affected by delayed consequences. Unless otherwise noted, we use a seven-day follow-up window and a three-day lookback window for high-demand coverage.

\paragraph{Stage 1: Product candidate selection.}
This stage measures whether the agent allocates attention to a sufficiently broad and relevant product set. From the tool traces, we extract executed state-changing product actions, including replenishment orders and price updates. Let $\mathcal{U}_{t}$ be the set of products receiving at least one such action on day $t$. Let $\mathcal{H}_{t}$ be the retrospective high-demand set, formed by the top-10 realized-sales products on day $t$ together with products that experienced stockout on that day. We report
\begin{equation}
\mathrm{Acted/day}
=\frac{1}{T}\sum_{t=1}^{T}|\mathcal{U}_{t}|,
\end{equation}
\begin{equation}
\mathrm{HighDemandCov}
=1-\frac{\sum_t|\mathcal{H}_{t}\setminus\mathcal{U}_{t-w:t}|}
{\sum_t|\mathcal{H}_{t}|},
\end{equation}
where $\mathcal{U}_{t-w:t}=\bigcup_{\tau=\max(1,t-w)}^{t}\mathcal{U}_{\tau}$ and $w=3$. \textbf{Acted/day} captures action breadth. \textbf{HighDemandCov} captures whether high-demand or stockout-affected products received recent operational attention. Auxiliary diagnostics include the number of distinct acted products, category coverage, action concentration, the top-10 product action share, and whether an acted product sells within seven days. These auxiliary quantities help distinguish a genuinely broad policy from a narrow policy that repeatedly acts on only a few products.

\paragraph{Stage 2: evidence acquisition.}
This stage measures whether a product action is supported by relevant same-day observations. Let $\mathcal{B}$ denote the evaluated executed product actions. For each action $a\in\mathcal{B}$, we define a required evidence set $\mathcal{R}(a)$. Replenishment orders require inventory, recent sales, supplier prices, and supplier-quality proxies such as customer ratings or supplier return rates. Price updates require current price, cost, inventory, and recent sales/profit history. Let $\mathcal{Q}(a)$ be the evidence categories queried earlier on the same day before $a$, and let $m_a=1$ if at least one required category is missing. We report
\begin{equation}
\mathrm{QDepth}
=\frac{1}{|\mathcal{B}|}\sum_{a\in\mathcal{B}}
\frac{|\mathcal{Q}(a)\cap\mathcal{R}(a)|}{|\mathcal{R}(a)|}.
\end{equation}
\begin{equation}
\mathrm{Completeness}
=1-\frac{1}{|\mathcal{B}|}\sum_{a\in\mathcal{B}}m_a.
\end{equation}
\textbf{QDepth} measures partial evidence coverage, while \textbf{Completeness} requires all action-critical evidence categories to be present. We also record the average gap between the last relevant query and the action, the number of distinct evidence tools used before an action, and same-day query--action overlap. For the oracle policy, \textbf{QDepth} and \textbf{Completeness} are set to 1 because the policy is a structured rule system whose inputs explicitly include the required inventory, demand, supplier, and price fields.

\paragraph{Stage 3: action conversion.}
This stage measures whether evidence is converted into economically meaningful actions. For price actions, let $p_a$ be the selected price and $p_a^\star$ be the historical-profit optimum estimated from visible sales and cost records. We first compute the absolute percentage distance
\begin{equation}
d_a=100\cdot\frac{|p_a-p_a^\star|}{p_a^\star}.
\end{equation}
The text reports the run-level mean distance $\bar d_r$ directly. For the normalized diagnostic plot in Figure~\ref{fig:failure_diagnostics}(b), we convert distance into an upper-right-is-better run-level score:
\begin{equation}
\mathrm{PriceClose}_r=1-\frac{\bar d_r}{\max_{r'}\bar d_{r'}},
\end{equation}
where the maximum is taken over valid plotted runs. Thus larger values indicate prices closer to the historical-profit optimum.

For supplier actions, let $r_a^{q}$ be the rank of the chosen supplier by raw quality among $K(a)$ same-product candidates, where rank 1 is best. We report
\begin{equation}
\mathrm{SupplierQual}(a)=
\frac{K(a)-r_a^{q}}{K(a)-1}.
\end{equation}
Because each product has five suppliers in our environment, this score maps the best-quality supplier to 1 and the worst-quality supplier to 0. We additionally report \textbf{PriceFirst}, the fraction of order lines choosing the cheapest supplier, and \textbf{QualityFirst}, the fraction choosing the raw-quality-best supplier. These two rates are post-hoc diagnostics: raw supplier quality is part of the simulator state and is not directly exposed to LLM agents, although agents can observe proxies such as return rates and customer ratings. A high \textbf{PriceFirst} rate with a low \textbf{QualityFirst} rate indicates a low-cost sourcing tendency that may increase returns and harm future reviews.

\begin{figure}[t]
\centering
\includegraphics[width=\linewidth]{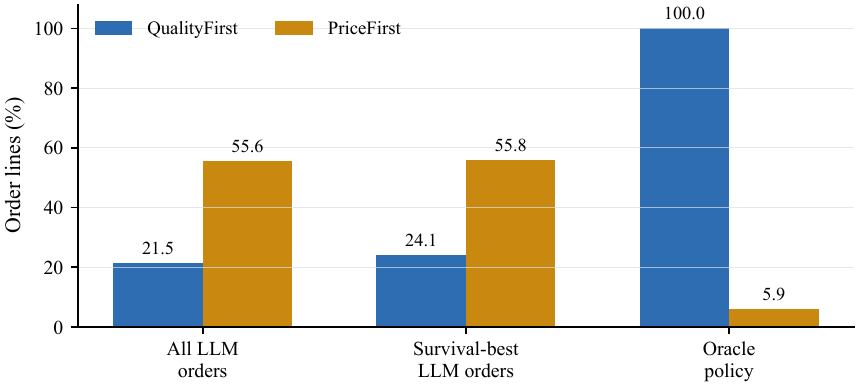}
\caption{Supplier-selection diagnostic used in Stage 3. \textbf{QualityFirst} is the fraction of order lines choosing the raw-quality-best supplier among candidates for the same product, while \textbf{PriceFirst} is the fraction choosing the cheapest supplier. The comparison shows that LLM orders are more price-first than quality-first, whereas the oracle policy explicitly prioritizes supplier quality.}
\label{fig:supplier_selection_diagnostic}
\end{figure}

\paragraph{Stage 4: temporal follow-up.}
This stage measures whether the agent maintains stateful attention after actions and delayed consequences. Let $\mathcal{P}$ be the set of acted product-day pairs. For each $(j,t)\in\mathcal{P}$, let $b_{jt}=1$ if product $j$ receives another state-changing action within seven days, and otherwise $b_{jt}=0$. We also collect delayed event records $\mathcal{D}$, including stockout, return, and expiration events. Let $u_e=1$ if event $e\in\mathcal{D}$ receives no later query or action within the seven-day analysis window, and otherwise $u_e=0$. We report
\begin{equation}
\mathrm{FollowUp}
=\frac{1}{|\mathcal{P}|}\sum_{(j,t)\in\mathcal{P}} b_{jt},
\end{equation}
\begin{equation}
\mathrm{Resolved}
=1-\frac{1}{|\mathcal{D}|}\sum_{e\in\mathcal{D}}u_e.
\end{equation}
\textbf{FollowUp} captures whether products that received actions remain in the operational loop. \textbf{Resolved} captures whether delayed negative signals are later attended to through either a query or another state-changing action. Additional diagnostics track later query/action revisits, response latency in days, repeated events without intervention, and adjacent-day action-set continuity. Together, these metrics distinguish a one-shot daily policy from a policy that maintains product-level state across delayed feedback.

\end{document}